\documentclass[journal]{IEEEtran}
\usepackage{amsmath,amsfonts}
\usepackage{algorithmic}
\usepackage{array}
\usepackage[caption=false,font=normalsize,labelfont=sf,textfont=sf]{subfig}
\usepackage{textcomp}
\usepackage{stfloats}
\usepackage{url}
\usepackage{verbatim}
\usepackage{graphicx}

\def\BibTeX{{\rm B\kern-.05em{\sc i\kern-.025em b}\kern-.08em
    T\kern-.1667em\lower.7ex\hbox{E}\kern-.125emX}}
\usepackage{balance}

\usepackage{booktabs,makecell, multirow, tabularx}
\usepackage{microtype}		
\usepackage{hyperref}
\hypersetup{
    colorlinks=true,
    linkcolor=blue,
    citecolor=green,
    urlcolor=magenta,
    filecolor=cyan,
}

\usepackage{cleveref} 		
\crefname{figure}{Fig.}{Figs.}
\crefname{table}{Table}{Tables}
\crefname{equation}{Eq.}{Eqs.}

\newcommand{\eg}{\textit{e.g.}}
\newcommand{\ie}{\textit{i.e.}}

\usepackage{amssymb}
\usepackage{pifont}
\newcommand{\cmark}{\ding{51}} 

\begin{document}

\title{Multiscale Switch for Semi-Supervised and Contrastive Learning in Medical Ultrasound Image Segmentation}

\author{
	Jingguo Qu, Xinyang Han, Yao Pu, Man-Lik Chui, Simon Takadiyi Gunda, Ziman Chen, Jing Qin \IEEEmembership{Senior Member, IEEE}, Ann Dorothy King, Winnie Chiu-Wing Chu, Jing Cai \IEEEmembership{Member, IEEE}, and Michael Tin-Cheung Ying
	\thanks{Received 25 December 2024; revised 1 September 2025 and 24 November 2025; accepted 24 February 2026. This work was supported by the General Research Funds of the Research Grant Council of Hong Kong under Grant 15102222 and Grant 15102524. \textit{(Corresponding author: Michael Tin-Cheung Ying.)}}
	\thanks{Jingguo Qu, Xinyang Han, Yao Pu, Man-Lik Chui, Simon Takadiyi Gunda, Ziman Chen, Jing Cai, and Michael Tin-Cheung Ying are with the Department of Health Technology and Informatics, The Hong Kong Polytechnic University, Hong Kong, China (e-mail: michael.ying@polyu.edu.hk).}
	\thanks{Jing Qin is with the Centre for Smart Health and School of Nursing, The Hong Kong Polytechnic University, Hong Kong, China.}
	\thanks{Ann Dorothy King and Winnie Chiu-Wing Chu are with the Department of Imaging and Interventional Radiology, The Chinese University of Hong Kong, Hong Kong, China.}
	\thanks{Digital Object Identifier 10.1109/TNNLS.2026.3669814}
}

\markboth{IEEE Transactions on Neural Networks and Learning Systems, Early Access}{Multiscale Switch for Semi-Supervised and Contrastive Learning in Medical Ultrasound Image Segmentation}

\maketitle

\begin{abstract}
	Medical ultrasound image segmentation faces significant challenges due to limited labeled data and characteristic imaging artifacts including speckle noise and low-contrast boundaries. While semi-supervised learning (SSL) approaches have emerged to address data scarcity, existing methods suffer from suboptimal unlabeled data utilization and lack robust feature representation mechanisms. In this paper, we propose Switch, a novel SSL framework with two key innovations: (1) Multiscale Switch (MSS) strategy that employs hierarchical patch mixing to achieve uniform spatial coverage; (2) Frequency Domain Switch (FDS) with contrastive learning that performs amplitude switching in Fourier space for robust feature representations. Our framework integrates these components within a teacher-student architecture to effectively leverage both labeled and unlabeled data. Comprehensive evaluation across six diverse ultrasound datasets (lymph nodes, breast lesions, thyroid nodules, and prostate) demonstrates consistent superiority over state-of-the-art methods. At 5\% labeling ratio, Switch achieves remarkable improvements: 80.04\% Dice on LN-INT, 85.52\% Dice on DDTI, and 83.48\% Dice on Prostate datasets, with our semi-supervised approach even exceeding fully supervised baselines. The method maintains parameter efficiency (1.8M parameters) while delivering superior performance, validating its effectiveness for resource-constrained medical imaging applications. The source code is publicly available at \url{https://github.com/jinggqu/Switch}.
\end{abstract}


\section{Introduction}

Medical imaging is an important technique in clinical diagnostics, providing the advantage of non-invasive visualization of internal body structures. Common techniques include computed tomography (CT), magnetic resonance imaging (MRI), and ultrasound (US) imaging. US is commonly utilized for diagnosing superficial organs and tissues, such as breast lesions~\cite{screening_brem_2015}, thyroid nodules~\cite{thyroid_takashima_1995}, and lymph nodes~\cite{sonography_ying_2003, sonographic_ahuja_2005, ultrasound_ahuja_2008} because of its real-time imaging capability, non-invasiveness, and cost-effectiveness. By employing high-frequency US waves, US generates detailed images of internal structures and is widely adopted in clinical settings. The current identification process of regions of interest (ROIs) in US images is performed manually by radiologists, which is time-consuming and prone to variations in accuracy and consistency, depending on the expertise of the radiologists.

Segmentation of US images presents significant challenges due to three primary factors. First, the quality of US images is often lower than CT and MRI because of the presence of speckle noise, which is a granular pattern that degrades image quality. More specifically, the ambiguity of ROI boundaries contributes to low contrast and artifacts caused by noise in US images. Second, the shape, size, and position of ROIs may vary significantly across images obtained from different scan planes and different patients. Third, US images frequently exhibit inconsistent brightness, resolution, and quality, which are influenced by variations in imaging settings and operator practices. These factors make the manual annotation process labor-intensive and time-consuming, highlighting the necessity for automated systems that are capable of accurately identifying ROIs to enhance both the accuracy and efficiency of diagnostic workflows.

\begin{figure}[t]
	\centering
	\includegraphics[width=0.85\linewidth]{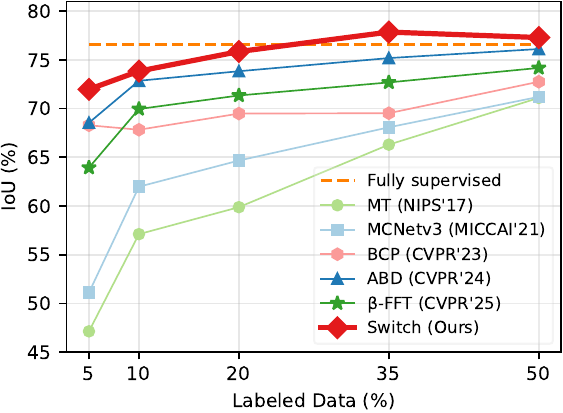}
	\caption{A comparison of model performance between previous semi-supervised SOTAs and the proposed Switch method on the lymph node dataset (LN-INT). Our method achieved the best results across five data labeling ratios (i.e. models were trained using only part of the ground truth).}
	\label{fig:ln-int}
\end{figure}

Deep learning has achieved remarkable success in image and signal processing tasks~\cite{learning_li_2020,multilevel_zhang_2020,multiview_cui_2024,infrared_khan_2024}. Building upon these advances, researchers have increasingly applied deep learning to medical image processing~\cite{unet_ronneberger_2015,retinal_zhao_2020,application_qu_2025,ai_han_2025,adapting_qu_2025}, which typically requires a substantial amount of well-annotated data. This rigorous requirement may be alleviated by the advent of semi-supervised learning (SSL)~\cite{semi_van_2020}. SSL models effectively leverage a limited amount of labeled data in conjunction with a vast quantity of unlabeled data to learn representations while maintaining coherence. This approach offers a more viable solution for medical image segmentation. Despite numerous SSL studies aiming to mitigate this constraint, the application of SSL in US image segmentation remains relatively limited in scope.

Current SSL studies of US image segmentation mainly focus on specific lesion locations, such as breast~\cite{semi_han_2020,semi_zhai_2022,residual_farooq_2023,ph_jiang_2024} and thyroid~\cite{an_wang_2019,deep_chen_2023}. Attention-based generative adversarial networks (GANs)~\cite{semi_han_2020,semi_zhai_2022} are designed to handle individual variance in breast lesions and improve the distinction between lesions and background through the discriminator of GANs. PH-Net~\cite{ph_jiang_2024} partitions the input image into multiple equal-size patches, introducing adaptive patch augmentation and hard patch shielding strategies with high average entropy for further model training. SABR-Net~\cite{deep_chen_2023} proposes a boundary refinement module to tackle the challenge of unclear edges within US images while introducing computational complexity. Although the aforementioned methods exhibit superior performance compared to state-of-the-art (SOTA) approaches across various US datasets, their generalizability remains uncertain due to the absence of thorough validation on other US datasets with comparable characteristics.

To address this issue, this study proposes a simple and effective SSL framework (named \textbf{Switch}), which is based on the teacher-student model~\cite{meanteacher_tarvainen_2017} for superficial US image segmentation. Our approach integrates three key components: multiscale switch (MSS), frequency domain switch (FDS), and contrastive learning (CL) modules. First, a pair of US images with or without manual annotations are partially switched by MSS to integrate coarse and fine representations. Second, the frequency domains of the above images decomposed by Fourier transformation are partially exchanged to reconstruct images that incorporate the frequency information of unlabeled images. This process generates positive and negative sample pairs for CL. Finally, the CL module is employed to reinforce the representation of the student network by maximizing the coherence between positive pairs in feature space, and vice versa. The proposed framework is evaluated on six superficial US datasets: the in-house lymph node US datasets (LN-INT and LN-EXT), the breast US image segmentation (BUSI) dataset~\cite{busi_al_2020}, thyroid datasets DDTI~\cite{ddti_pedraza_2015} and TN3K~\cite{tn3k_gong_2023}, and the Prostate~\cite{microsegnet_jiang_2023} dataset. As shown in~\Cref{fig:ln-int}, the proposed Switch consistently outperformed previous SOTAs on the LN-INT dataset across five labeling ratios, where each ratio denotes the percentage of annotated samples used for training. The key contributions of this study are outlined as follows:
\begin{itemize}
	\item We propose a novel SSL framework for superficial US image segmentation with similar essence, which fuse the coarse and fine knowledge to strengthen the consistency within image pairs.
	\item We develop a CL module with frequency domain switch strategy to boost the uniformity between the original and reconstructed images.
	\item We conduct extensive experiments on six superficial US datasets with an external testing set and demonstrate the effectiveness and generalizability of the proposed framework.
\end{itemize}

The remainder of this paper is organized as follows: \Cref{sec:related-work} reviews the related work on medical image segmentation, SSL, and CL. \Cref{sec:method} presents the detailed methodology of the proposed Switch framework, including the MSS, the FDS for CL, consistency regularization, augmentations, loss function, and training strategy. \Cref{sec:exp} describes the experimental setup, datasets, evaluation metrics, implementation details, and provides comprehensive comparisons with SOTA methods along with ablation studies. \Cref{sec:discussion} provides comprehensive ablation studies, embedding space analysis, and discusses the clinical implications and limitations of our approach. Finally, \Cref{sec:conclusion} concludes the paper.
\section{Related Work}
\label{sec:related-work}

\subsection{Medical Image Segmentation}

Medical image segmentation refers to the classification process of 3D volumes or 2D images to extract ROIs at the pixel level, which is a fundamental task in medical image analysis and clinical applications. Many deep learning-based methods have been proposed for this task, including U-Net~\cite{unet_ronneberger_2015} and its variants~\cite{vnet_milletari_2016, unetplusplus_zhou_2018, 3dunet_cicek_2016}. Other methods have also achieved SOTA performance as the backbone of segmentation networks, such as DeepLab~\cite{deeplab_chen_2018}, PSPNet~\cite{pspnet_zhao_2017}, and HRNet~\cite{hrnet_wang_2021}.

\subsection{Semi-Supervised Learning}

SSL methods aim to learn global representations across the entire dataset by utilizing both labeled and unlabeled data. There are two main categories of SSL methods: consistency regularization and pseudo labeling. Consistency regularization methods attempt to minimize the difference between predictions on the same input with different augmentations or views, while pseudo labeling methods aim to generate high-quality pseudo labels for unlabeled data and combine them with labeled data to strengthen the model. Many approaches based on consistency regularization~\cite{meanteacher_tarvainen_2017, cutmix_yun_2019, uamt_yu_2019, cct_ouali_2020, classmix_olsson_2021, mcnet_wu_2022, ugmcl_zhang_2023, augmentation_zhao_2023, bcp_bai_2023, abd_chi_2024} and pseudo labeling~\cite{curriculum_cascante_2021, stplusplus_yang_2022, semi_wang_2022, semisam_deng_2024} have been proposed to address SSL. Particularly, Mean Teacher~\cite{meanteacher_tarvainen_2017} enhances the performance of the student network by enforcing coherence between the predictions of student and teacher networks through an exponential moving average (EMA) strategy. UA-MT~\cite{uamt_yu_2019} introduces Monte Carlo dropout to estimate and minimize the uncertainty between the student and teacher networks. CCT~\cite{cct_ouali_2020} applies multiple auxiliary segmentation heads with randomly perturbed encoded features to reduce the discrepancy between these two networks. Copy-Paste (CP)~\cite{cutmix_yun_2019} design has also been utilized for SSL, where BCP~\cite{bcp_bai_2023} adopts a bidirectional CP approach to further increase the uniformity within the dataset. In addition, large-scale pre-trained vision models such as the Segment Anything Model (SAM)~\cite{sam_kirillov_2023} have been incorporated to generate high-quality pseudo labels for unlabeled samples~\cite{semisam_deng_2024}. Recent advances include PH-Net~\cite{phnet_jiang_2024} which introduces patch-wise hardness estimation for breast lesion segmentation, and ABD~\cite{abd_chi_2024} which proposes adaptive bidirectional displacement strategies. Furthermore, $\beta$-FFT~\cite{betafft_hu_2025} presents nonlinear interpolation in frequency domain for enhanced training strategies.

\subsection{Contrastive Learning}

CL methods were proposed to learn global representations in a self-supervised manner by maximizing and minimizing the similarity between positive and negative pairs, respectively~\cite{dimensionality_hadsell_2006, instdisc_wu_2018, simclr_chen_2020, moco_he_2020, exploring_wang_2021}. Many efforts have been made to improve the performance of SSL by incorporating CL. CDCL~\cite{crosspatch_wu_2022} constructs negative pairs from feature patches with large disparity to enhance the discrimination capability of the segmentation model. MMS~\cite{mms_lou_2023} introduces independent classifiers and projectors to conduct supervised and unsupervised CL in feature space according to spatial correspondence at the pixel level. In addition, U$^2$PL~\cite{semi_wang_2022} filters out unreliable pseudo labels predicted by the teacher model through entropy estimation as negative samples, and then these negative samples are pushed into a memory bank~\cite{instdisc_wu_2018} to provide consistent and continually updated support for the student model. Similarly, PH-Net~\cite{phnet_jiang_2024} shields patches with high entropy to avoid them being altered through CutMix~\cite{cutmix_yun_2019} operations at the patch level. A separate projector and memory bank are also associated with the high reliability patch selection, sampling, and CL processes. These current CL methods for semantic segmentation mainly focus on the overall variability between labeled and unlabeled data, while ignoring the information cohesion between them. Following this observation, we propose a novel sample pair construction method for CL by employing frequency domain switching to enhance the local harmony and uniformity within US image pairs.

\subsection{Summary}

Despite promising results in general medical image segmentation, existing SSL and CL methods face two critical limitations for US image segmentation: (1) \textit{Geometric Adaptation Challenges}: Fixed-patch methods like Copy-Paste~\cite{copypaste_ghiasi_2021} and BCP~\cite{bcp_bai_2023} inadequately handle US ROIs with highly variable shapes and positions across scan planes; (2) \textit{Feature Representation Limitations}: Current CL approaches emphasize overall variability while neglecting information cohesion between labeled and unlabeled data. To address these limitations, we propose Switch which integrates multiscale spatial switching with frequency domain manipulation for enhanced US image segmentation.  The detailed methodology is presented in the following section.

\section{Method}
\label{sec:method}

\begin{figure*}[htb]
	\centering
	\includegraphics[width=1\linewidth]{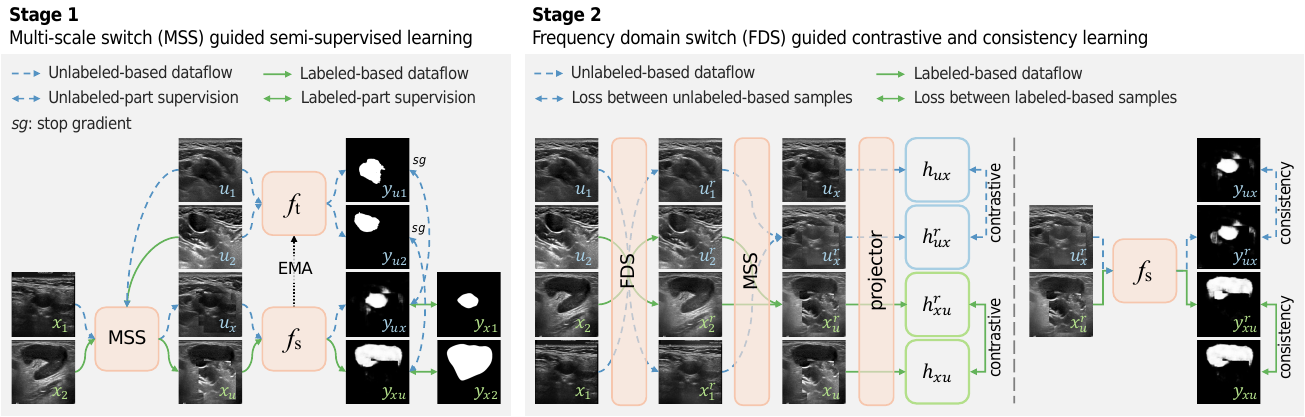}
	\caption{Overview of our proposed method. $f_\mathrm{s}$ and $f_\mathrm{t}$ are the student and teacher network, respectively. \textbf{Stage 1}: The system splits labeled data into two subsets ($x_1$, $x_2$) and unlabeled data into two subsets ($u_1$, $u_2$), generates pseudo-labels for unlabeled images using $f_\mathrm{t}$, creates spatial masks for patch-based mixing, and combines unlabeled base images with labeled patches to create mixed training samples for semi-supervised learning. \textbf{Stage 2}: The system performs amplitude switching between labeled and unlabeled image batches in the frequency domain, creates new mixed images using the same spatial masks from MSS, and applies CL to align feature representations between original and frequency-switched mixed images, while enforcing consistency regularization between their model predictions to enhance semi-supervised learning robustness. More details of MSS and FDS modules can be found in~\Cref{fig:mss} and~\Cref{fig:fds}.}
	\label{fig:overview}
\end{figure*}

The overall structure of our proposed method is shown in~\Cref{fig:overview}, which is based on the Mean Teacher framework~\cite{meanteacher_tarvainen_2017}. Our approach employs a teacher-student architecture consisting of two neural networks with identical U-Net~\cite{unet_ronneberger_2015} architectures that collaborate to leverage both labeled and unlabeled data effectively.

In this framework, the student network $f_\mathrm{s}$ serves as the primary learning model that is actively trained using gradient descent on both labeled and unlabeled data. It receives mixed samples generated through our multiscale switch mechanism and learns to predict segmentation masks under supervision from both ground truth labels and pseudo labels. The teacher network $f_\mathrm{t}$, in contrast, acts as a stable reference model that generates reliable pseudo labels for unlabeled data. Crucially, the teacher network is not directly trained through gradient descent; instead, it maintains a temporally averaged version of parameters derived from the student network through exponential moving average (EMA) updates. The EMA update mechanism can be mathematically expressed as:
\begin{equation}
	\theta_\mathrm{t}^{(k+1)} = \alpha \theta_\mathrm{t}^{(k)} + (1-\alpha) \theta_\mathrm{s}^{(k+1)}
	\label{eq:ema_update}
\end{equation}
where $\theta_\mathrm{t}^{(k)}$ and $\theta_\mathrm{s}^{(k)}$ represent the parameters of teacher and student networks at iteration $k$, respectively, and $\alpha$ is the momentum coefficient (typically set to 0.99).

For ease of description, we hereby make the following mathematical definitions.

Given a medical US dataset $\mathcal{D} = \{\mathcal{D}_l \cup \mathcal{D}_u\}$, $\mathcal{D}_l = \{x_i, y_i\}_{i=1}^{N}$ consists of $N$  images with ground truth annotations and $\mathcal{D}_u = \{u_i\}_{i=1}^{M}$ only includes $M$ ($N \ll M$) unlabeled images, where $x_i$, $u_i$ and $y_i$ are the input image with and without annotation and corresponding ground truth label, respectively.

\subsection{Multiscale Switch}\label{sec:mss}

In CP series methods~\cite{cutmix_yun_2019, bcp_bai_2023}, only fixed-area arbitrary regions of input images were used to fuse sample pairs as input to the student model, which may lack the concentration for US images with variable ROI size and location. To address this issue and inspired by SwAV~\cite{swav_caron_2020} and BCP~\cite{bcp_bai_2023}, we propose an MSS mechanism to incorporate partial unlabeled information into labeled samples.

First, a binary mask $\mathcal{M}$ consisting of $p$ coarse patches and $q$ fine patches is randomly generated for sample batches. The mask generation algorithm can be formalized as:
\begin{equation}
	\mathcal{M} = \bigcup_{i=1}^{p} \mathcal{P}_\mathrm{coarse}^{(i)} \cup \bigcup_{j=1}^{q} \mathcal{P}_\mathrm{fine}^{(j)}
	\label{eq:mask_generation}
\end{equation}
where $\mathcal{P}_\mathrm{coarse}^{(i)}$ and $\mathcal{P}_\mathrm{fine}^{(j)}$ represent the $i$-th coarse patch (size $s_c \times s_c = 128 \times 128$) and $j$-th fine patch (size $s_f \times s_f = 32 \times 32$), respectively. Each patch is positioned according to:
\begin{equation}
	\begin{aligned}
		\mathcal{P}_\mathrm{coarse}^{(i)} & = \{(x,y) : x_i \leq x < x_i + s_c, y_i \leq y < y_i + s_c\} \\
		\mathcal{P}_\mathrm{fine}^{(j)}   & = \{(x,y) : x_j \leq x < x_j + s_f, y_j \leq y < y_j + s_f\}
	\end{aligned}
	\label{eq:patch_coordinates}
\end{equation}
where $(x_i, y_i)$ and $(x_j, y_j)$ are randomly sampled upper-left coordinates satisfying $x_i \sim \mathcal{U}(0, H-s_c)$, $y_i \sim \mathcal{U}(0, W-s_c)$, $x_j \sim \mathcal{U}(0, H-s_f)$, and $y_j \sim \mathcal{U}(0, W-s_f)$ for image dimensions $H \times W$, where $\mathcal{U}(a, b)$ denotes the uniform distribution over the interval $[a, b)$.

\begin{figure}[t]
	\centering
	\includegraphics[width=0.9\linewidth]{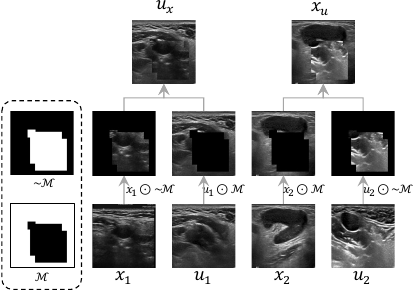}
	\caption{Overview of multiscale switch mechanism. $\mathcal{M}$ represents the binary mask with $p$ coarse patches and $q$ fine patches, while $\mathbin{\sim}\mathcal{M}$ refers to its complement. The white areas in both $\mathcal{M}$ and $\mathbin{\sim}\mathcal{M}$ stand for True, and the pixel at the corresponding position will be retained after multiplication. Vice versa for black regions.}
	\label{fig:mss}
\end{figure}

Second, the labeled and unlabeled sample pair is randomly selected from $\mathcal{D}_l$ and $\mathcal{D}_u$ to conduct the MSS. This process can be seen in~\Cref{fig:mss} and described mathematically as follows:
\begin{equation}
	\begin{aligned}
		u_x & = u_1 \odot \mathcal{M} + x_1 \odot \mathbin{\sim} \mathcal{M} \\
		x_u & = x_2 \odot \mathcal{M} + u_2 \odot \mathbin{\sim} \mathcal{M} \\
	\end{aligned}
\end{equation}
where $\odot$ denotes element-wise multiplication, $\mathbin{\sim} \mathcal{M}$ is the complementary mask of $\mathcal{M}$, $x_1, x_2 \in \mathcal{D}_l$, $u_1, u_2 \in \mathcal{D}_u$, and $u_x$ and $x_u$ are a sample pair obtained after MSS operation and both contain parts of the image from each other.

Finally, the predicted labels $y_{ux}$ and $y_{xu}$ originated from the reassembled samples $u_x$ and $x_u$ are collected from the student network. The pseudo labels $y_{u1}$ and $y_{u2}$ corresponding to the unlabeled images $u_1$ and $u_2$ are yielded from the teacher network simultaneously. The pseudo label generation process can be expressed as:
\begin{equation}
	y_{ui} = \arg\max(\sigma(f_\mathrm{t}(u_i)))
	\label{eq:pseudo_label}
\end{equation}
where $\sigma(\cdot)$ denotes the softmax function. To improve pseudo label quality, we apply connected component analysis to retain only the largest connected component:
\begin{equation}
	y_{ui} = \mathrm{LCC}(y_{ui})
	\label{eq:pseudo_label_nms}
\end{equation}
where $\mathrm{LCC}(\cdot)$ represents the largest connected component operation.

Mixed Dice loss and cross-entropy loss are calculated with region-specific weights. For the mixed sample $u_x$ (unlabeled as base), the loss is formulated as:
\begin{equation}
	\begin{aligned}
		\mathcal{L}_{ux} & = w_b \mathcal{L}(y_{ux}, y_{u1}, \mathcal{M}) + w_p \mathcal{L}(y_{ux}, y_{x1}, \mathbin{\sim}\mathcal{M}) \\
		\mathcal{L}_{xu} & = w_b \mathcal{L}(y_{xu}, y_{x2}, \mathcal{M}) + w_p \mathcal{L}(y_{xu}, y_{u2}, \mathbin{\sim}\mathcal{M})
	\end{aligned}
	\label{eq:mixed_loss}
\end{equation}
where $w_b$ and $w_p$ are the base area weight and patch area weight, respectively, $y_{x1}$ and $y_{x2}$ are the ground truth labels for $x_1$ and $x_2$, and $\mathcal{L}(\cdot)$ represents the combination of Dice and cross-entropy losses. The Dice loss is defined as follows:
\begin{equation}
	\mathcal{L}_{\mathrm{dice}} = 1 - \frac{2 \sum_i p_i g_i + \epsilon}{\sum_i p_i + \sum_i g_i + \epsilon}
	\label{eq:dice_loss}
\end{equation}
where $p$ and $g$ refer to prediction and ground truth,  while the inclusion of a small smoothing factor $\epsilon$ ensures numerical stability.

\subsection{Frequency Domain Switch}

\begin{figure}[t]
	\centering
	\includegraphics[width=0.9\linewidth]{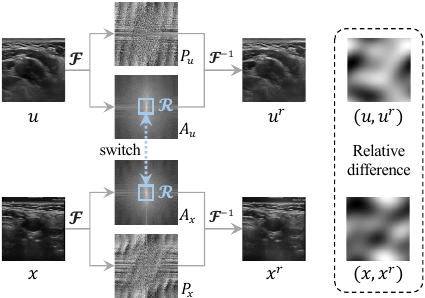}
	\caption{Overview of frequency domain switch approach. $x$ and $u$ are decomposed into amplitude ($A_x, A_u$) and phase ($P_x, P_u$) domain by fast Fourier transform ($\mathcal{F}$). The frequency region ($\mathcal{R}$) in amplitude domain is switched between $A_x$ and $A_u$, and then the pair of $x^r$ and $u^r$ are reconstructed by inverse fast Fourier transform ($\mathcal{F}^{-1}$). The relative difference of ($x, x^r$) and ($u, u^r$) made by FDS is shown in the dotted box.}
	\label{fig:fds}
\end{figure}

In previous CL methods~\cite{swav_caron_2020, simclr_chen_2020, mms_lou_2023}, the construction process of contrasting sample pairs is primarily based on the same image with different types of augmentations. This procedure may impair the original information contained in raw input data and ignores the relationship between labeled and unlabeled data. Unlike frequency domain adaptation (FDA)~\cite{fda_yang_2020} which transfers style information between different domains for domain adaptation, we propose an FDS approach for cross-sample frequency mixing within the same ultrasound domain to enhance labeled-unlabeled data relationships for SSL. The whole process is shown in~\Cref{fig:fds}.

The overall style and pattern of an image are mainly stored in the low-frequency area, while the high-frequency area contains information about drastically shifting details, such as edges~\cite{fda_yang_2020, feddg_liu_2021}, especially speckle noise in US images. Different from FDA that manipulates broader low-frequency regions for cross-domain style transfer, our FDS performs cross-sample amplitude mixing in a carefully controlled small frequency region to preserve anatomical structure while enabling texture information exchange for SSL.

The FDS execution involves four sequential steps:

(1) \textit{Frequency Decomposition}. Specifically, the input pair $x$ and $u$ are decomposed into amplitude and phase domains by fast Fourier transform ($\mathcal{F}$) with zero-frequency shifting to the center:
\begin{equation}
	\begin{aligned}
		F_x & = \mathrm{fftshift}(\mathcal{F}(x)) = \mathrm{fftshift}(A_x e^{i P_x}) \\
		F_u & = \mathrm{fftshift}(\mathcal{F}(u)) = \mathrm{fftshift}(A_u e^{i P_u})
	\end{aligned}
	\label{eq:ft}
\end{equation}
where $A_x = |F_x|$, $P_x = \angle F_x$, $A_u = |F_u|$, and $P_u = \angle F_u$ represent the amplitude and phase components, respectively.

(2) \textit{Frequency Filtering}. We define a centralized low-frequency region $\mathcal{R}$ to control the extent of information exchange. This region is formulated as a square mask centered at the zero-frequency component:
\begin{equation}
	\mathcal{R} = \{(i,j) : |i - H/2| \leq r_H, |j - W/2| \leq r_W\}
	\label{eq:freq_region}
\end{equation}
where $r_H = \lfloor H \cdot \rho \rfloor / 2$ and $r_W = \lfloor W \cdot \rho \rfloor / 2$. The parameter $\rho$ (frequency area ratio, typically 0.0175) is a critical hyperparameter that balances diversity and realism. A small $\rho$ targets only the global style components, while a larger $\rho$ would introduce excessive high-frequency noise exchange that could degrade structural integrity.

(3) \textit{Amplitude Switching}. The low-frequency amplitude components between $x$ and $u$ are exchanged by using the binary mask $\mathcal{R}$ and its complement $\mathbin{\sim} \mathcal{R}$:
\begin{equation}
	\begin{aligned}
		A^r_u & = A_u \odot \mathbin{\sim} \mathcal{R} + A_x \odot \mathcal{R} \\
		A^r_x & = A_x \odot \mathbin{\sim} \mathcal{R} + A_u \odot \mathcal{R}
	\end{aligned}
	\label{eq:amplitude_switch}
\end{equation}

This operation retains the high-frequency content of each original image while transferring the low-frequency style from its counterpart.

(4) \textit{Image Reconstruction}. Finally, the augmented images are reconstructed by combining the modified amplitude spectra with their original phase components via the inverse fast Fourier transform:
\begin{equation}
	\begin{aligned}
		u^r & = \mathrm{real}(\mathcal{F}^{-1}(\mathrm{ifftshift}(A^r_u e^{i P_u}))) \\
		x^r & = \mathrm{real}(\mathcal{F}^{-1}(\mathrm{ifftshift}(A^r_x e^{i P_x})))
	\end{aligned}
	\label{eq:ift}
\end{equation}

By strictly preserving the original phase information ($P_u$ and $P_x$), the reconstructed images $u^r$ and $x^r$ maintain perfect pixel-level alignment with their corresponding semantic labels (or pseudo-labels). This ensures that the supervision signals remain valid despite the significant appearance transformations.

Subsequently, the MSS operation described in \Cref{sec:mss} is applied to the FDS-augmented pair ($x^r$, $u^r$) to generate the final mixed samples ($u^r_x$, $x^r_u$) for CL. To extract features for the contrastive objective, we employ a dedicated projection head.

The projection head consists of a convolutional block for feature extraction, a $2\times2$ max pooling layer for spatial downsampling, followed by a second convolutional block and pooling layer. A final $1\times1$ convolution projects the features into the target embedding space.

The projector is randomly initialized and its gradients from the FDS-augmented branch ($h^r_{ux}$, $h^r_{xu}$) are not back-propagated to the main encoder during CL. The feature projections are computed as:
\begin{equation}
	\begin{aligned}
		h_u = \mathrm{proj}(f_\mathrm{s}(u_x)), \quad h^r_{ux} = \mathrm{proj}(f_\mathrm{s}(u^r_x)) \\
		h_x = \mathrm{proj}(f_\mathrm{s}(x_u)), \quad h^r_{xu} = \mathrm{proj}(f_\mathrm{s}(x^r_u))
	\end{aligned}
\end{equation}
where $f_\mathrm{s}$ denotes the student network and $\mathrm{proj}$ represents the projection head.

In the CL framework, feature pairs from the same spatial location are treated as positive pairs and vice versa. The objective is to maximize the similarity between positive pairs while minimizing it for negative pairs, thereby encouraging the model to learn robust, invariant feature representations.

\subsection{Consistency Regularization}

To further enhance the robustness of the model predictions, we introduce a consistency regularization term that enforces the model to produce similar outputs for the original mixed images and their frequency-domain reconstructed counterparts. The consistency loss is formulated as:
\begin{equation}
	\mathcal{L}_\mathrm{consist} = \frac{1}{2}\left(\mathrm{MSE}(f_\mathrm{s}(u_x), f_\mathrm{s}(u^r_x)) + \mathrm{MSE}(f_\mathrm{s}(x_u), f_\mathrm{s}(x^r_u))\right)
	\label{eq:consistency_loss}
\end{equation}
where $\mathrm{MSE}(\cdot, \cdot)$ denotes the mean squared error between the logit outputs before softmax activation. This consistency constraint encourages the model to be invariant to frequency domain perturbations, thereby improving generalization capability.

\subsection{Augmentations}

The existing literature~\cite{cutmix_yun_2019, bcp_bai_2023, abd_chi_2024, semisam_deng_2024,phnet_jiang_2024} on medical image segmentation has not addressed the potential benefits of image augmentation for enhancing the robustness of proposed models, while some studies show that data augmentations can improve the performance of segmentation models~\cite{augmentation_zhao_2023}.

To address this issue, we apply a series of data augmentation strategies to enhance the robustness of the model. Following~\cite{augmentation_zhao_2023}, we divide the augmentation methods into two categories: weak and strong. The weak augmentation includes resize and crop, horizontal flip, and vertical flip, while the strong augmentation consists of auto contrast, Gaussian blur, contrast, brightness, sharpness, posterize, and solarize. The detailed explanation of each augmentation is shown in~\Cref{tab:augs}.

Unlike other medical imaging modalities, such as CT and MRI, the location and size of ROIs in US images are highly variable. Considering the fragile nature of US images, the intensity of each augmentation is reduced and augmentations with strong destructive effects on the entire image have also been removed, such as histogram equalization. A random amount of the augmentations in~\Cref{tab:augs} are applied to training data, and the parameters are randomly selected from the specified range.

\begin{table}[htb]
	\caption{Summary of augmentations.}
	\centering
	\footnotesize
	\setlength{\tabcolsep}{3pt}
	\begin{tabular}{lll}
		\toprule
		Type                    & Operation       & Explanation                                    \\
		\midrule
		None                    & Identity        & No augmentation                                \\
		\midrule
		\multirow{3}{*}{Weak}   & Resize and crop & Randomly resize the image by [0.8, 1.2]        \\
		                        & Horizontal flip & Horizontally flips the image                   \\
		                        & Vertical flip   & Vertically flips the image                     \\
		\midrule
		\multirow{7}{*}{Strong} & Auto contrast   & Maximizes (normalize) the image contrast       \\
		                        & Gaussian Blur   & Blurs the image with a Gaussian kernel         \\
		                        & Contrast        & Adjusts the contrast by [0.75, 1.25]           \\
		                        & Brightness      & Adjusts the brightness by [0.75, 1.25]         \\
		                        & Sharpness       & Adjusts the sharpness by [0.75, 1.25]          \\
		                        & Posterize       & Reduces each pixel to [4,8] bits               \\
		                        & Solarize        & Inverts pixels above a threshold from [1, 256) \\
		\bottomrule
	\end{tabular}
	\label{tab:augs}
\end{table}

\subsection{Loss Function}

There are three types of losses in our proposed method: the MSS loss, the contrastive loss for unsupervised learning, and the consistency regularization loss. The MSS loss is computed as the combination of Dice and cross-entropy losses for both mixed samples:
\begin{equation}
	\mathcal{L}_\mathrm{mss} = \frac{1}{4}(\mathcal{L}_{\mathrm{dice}}^{ux} + \mathcal{L}_{\mathrm{ce}}^{ux} + \mathcal{L}_{\mathrm{dice}}^{xu} + \mathcal{L}_{\mathrm{ce}}^{xu})
	\label{eq:mss_loss}
\end{equation}
where each component loss is computed using the region-specific weighting as described in~\Cref{eq:mixed_loss}.

To alleviate the complexity and computational cost of the contrastive loss, we combine the standard InfoNCE loss~\cite{infonce_oord_2018} with the projection outcomes $h^r_{ux}$ and $h^r_{xu}$ as the contrastive loss. FDS (\Cref{fig:fds}) alters low-frequency amplitude (texture/speckle) while preserving phase (anatomical structure), so positives capture the same anatomy under different texture/noise realizations. InfoNCE therefore encourages the encoder to learn structural content robustly and discard spurious speckle or intensity fluctuations. This complements Dice/CE, which supervise label prediction but do not directly enforce invariance or inter-class margins.

Given the batch size $B$ and the dimension of the feature map $K = H \times W$, there are $N = B \times K$ queries in total for CL. The contrastive loss is defined as:
\begin{equation}
	\begin{aligned}
		\mathcal{L}_\mathrm{cont} & = - \frac{1}{N} \sum_{b=1}^{B} \sum_{i=1}^{K} \log \frac{\exp\left( l_\mathrm{pos}^{b,i} / \tau \right)}{\sum_{j=0, j \neq i}^{K} \exp\left( l_\mathrm{neg}^{b,i,j} / \tau \right)} \\
		l_\mathrm{pos}^{b,i}      & = \bigl\{\langle h_{ux}^{b,i}, h_{ux}^{r, b,i}\rangle, \langle h_{xu}^{b,i}, h_{xu}^{r, b,i}\rangle\bigr\}                                                                          \\
		l_\mathrm{neg}^{b,i,j}    & = \bigl\{\langle h_{ux}^{b,i}, h_{ux}^{r, b,j} \rangle, \langle h_{xu}^{b,i}, h_{xu}^{r, b,j} \rangle, i \neq j\bigr\}
	\end{aligned}
\end{equation}
where $\tau$ is the temperature parameter to scale the similarity scores. $l_\mathrm{pos}^{b, i}$ refers to the similarity between the positive pair at spatial position $i$ in the $b$-th sample. $l_\mathrm{neg}^{b, i, j}$ denotes the similarity between the query feature $h_{ux}^{b, i}$ at position $i$ and the key feature $h_{ux}^{r, b, j}$ at position $i \neq j$ within the same sample. The same applies to the pair $h_{xu}^{b, i}$ and $h_{xu}^{r, b, j}$. The diagonal elements (when $i = j$) are set to negative infinity to avoid including self-similarities.

The final loss is a weighted linear combination of all three loss components:
\begin{equation}
	\mathcal{L}_\mathrm{total} = \mathcal{L}_\mathrm{mss} + \lambda_\mathrm{cont} \mathcal{L}_\mathrm{cont} + \lambda_\mathrm{consist} \mathcal{L}_\mathrm{consist}
	\label{eq:total_loss}
\end{equation}
where $\lambda_\mathrm{cont}$ and $\lambda_\mathrm{consist}$ represent the weights of contrastive loss and consistency loss, respectively.

\subsection{Training Strategy}

Our training strategy consists of two phases: pre-training and self-training. In the pre-training phase, only labeled data are used to train the student network $f_s$ using the MSS mechanism:
\begin{equation}
	\mathcal{L}_\mathrm{pre} = \frac{1}{2}(\mathcal{L}_\mathrm{dice} + \mathcal{L}_\mathrm{ce})
	\label{eq:pretrain_loss}
\end{equation}
This phase provides a strong initialization for the subsequent self-training phase.

In the self-training phase, both labeled and unlabeled data are utilized with the full loss function (\Cref{eq:total_loss}). The teacher network is initialized with the pre-trained student weights and updated using EMA (\Cref{eq:ema_update}). This two-phase training strategy ensures stable convergence and optimal utilization of both labeled and unlabeled data.

\section{Experiments}
\label{sec:exp}

The performance of the proposed method is evaluated on various US datasets and compared with SOTA methods.

\subsection{Datasets}

\textbf{Lymph Node (LN)}. We retrospectively collected two LN datasets from two different centers. \textbf{LN-INT} comprises 1,292 US images of 307 examinations from Prince of Wales Hospital of Hong Kong, with years ranging from 2016 to 2024, and \textbf{LN-EXT} includes 374 images. All images are annotated at the pixel level by experienced radiologists to delineate the LN region. Notably, LN-EXT is used exclusively as an external test set and is not involved in any part of the training process.

\textbf{BUSI}~\cite{busi_al_2020}. The BUSI dataset originally contains 780 breast lesion US images, with 437 benign, 210 malignant and 133 normal cases. The normal cases without ROIs are excluded in the following experiments as the purpose of this study is to detect and segment the abnormal regions, resulting in 647 images for analysis.

\textbf{DDTI}~\cite{ddti_pedraza_2015}. The DDTI dataset consists of 637 US images of thyroid nodules. The dataset includes a wide range of lesion types including focal thyroiditis, cystic nodules, adenomas and thyroid cancer.

\textbf{TN3K}~\cite{tn3k_gong_2023}. The TN3K dataset was collected at Zhujiang Hospital, South Medical University. A single representative image is retained among those captured from a similar viewpoint or within the same geographical area of a given patient. The TN3K dataset includes 2,879 and 614 images for training and testing, respectively.

\textbf{Prostate}~\cite{microsegnet_jiang_2023}. This dataset comprises both expert and non-expert prostate annotations of 75 patients who underwent micro-US-guided prostate biopsy, and only the expert annotations are used in this study. There are 2,152 images involved in the training process, while 758 images are only used for model testing.

The overview and partitions of the included datasets are shown in~\Cref{tab:datasets}. Datasets except TN3K~\cite{tn3k_gong_2023} are randomly divided into training, validation and test sets in the ratio of 8:1:1. It is worth noting that the pre-defined training set of TN3K~\cite{tn3k_gong_2023} is split into training and validation sets in an 8:2 ratio, while the test set remains unchanged.

\begin{table}[htb]
	\caption{Overview of datasets and partitions.}
	\centering
	\begin{tabular}{l|rrr|r}
		\toprule
		Dataset                                & Training & Validation & Testing & Total \\
		\midrule
		LN-INT                                 & 1,034    & 129        & 129     & 1,292 \\
		LN-EXT                                 & 0        & 0          & 374     & 374   \\
		BUSI~\cite{busi_al_2020}               & 518      & 65         & 64      & 647   \\
		DDTI~\cite{ddti_pedraza_2015}          & 510      & 64         & 63      & 637   \\
		TN3K~\cite{tn3k_gong_2023}             & 2,304    & 575        & 614     & 3,493 \\
		Prostate~\cite{microsegnet_jiang_2023} & 1,937    & 215        & 758     & 2,910 \\
		\bottomrule
	\end{tabular}
	\label{tab:datasets}
\end{table}

\subsection{Pre-processing}

For the private lymph node datasets, four pre-processing steps are applied to all US images to standardize the data and reduce irrelevant information. First, images are cropped by a specified margin (default 50 pixels) to remove device interfaces and extraneous boundary elements. Second, template matching with confidence threshold detection is employed to identify and eliminate fixed template regions such as device logos and measurement scales from the US images. Third, morphological operations and contour detection algorithms are utilized to automatically extract the largest connected US scanning region from the processed images. Fourth, the extracted US area is cropped inward by 2\% to remove residual artifacts and noise from the region boundaries.

All images in the above six datasets are then resized to 256$\times$256 pixels for training and testing.

\subsection{Evaluation Metrics}

We choose four widely-used metrics to evaluate the performance of the proposed method: Dice coefficient (Dice, \%), Jaccard index (intersection over union, IoU, \%), 95th-percentile of Hausdorff distance (HD95), and average symmetric surface distance (ASD). Dice coefficient and IoU quantify the volumetric overlap between predicted and ground-truth masks. The HD95 measures the maximum boundary discrepancy after excluding the largest 5\% of outliers, and ASD computes the mean distance between all corresponding boundary points in both directions. Higher Dice and IoU values, together with lower HD95 and ASD values, indicate superior segmentation performance. All metrics are calculated based on the binary segmentation results and averaged across all images in the test set.

\subsection{Implementation Details}

All experiments are conducted on the NVIDIA RTX 4090 GPU with a fixed random seed. The proposed method is implemented using PyTorch and optimized by the SGD optimizer with a learning rate of 0.05 that decreases by cosine annealing strategy. The student model is pre-trained for 10,000 iterations on the labeled dataset with MSS and the teacher model remains uninitialized during this process. Subsequently, training is performed on the entire dataset for 30,000 iterations, while the teacher model is updated by the student using EMA with momentum coefficient $\alpha = 0.99$ as described in~\Cref{eq:ema_update}. The sub-batch sizes for the labeled and unlabeled data are set to an equal value of 8 within each sample batch, with a total size of 16.

The amount of coarse patches $p$ and fine patches $q$ in MSS are set to 2 and 2 as default, respectively, with coarse patch size $s_c = 128$ and fine patch size $s_f = 32$. The frequency area ratio $\rho$ in FDS is set to 0.0175 (1.75\% of the image size). The region-specific weights in mixed loss computation are set as $w_b = 1.0$ (base area weight) and $w_p = 0.5$ (patch area weight). For the contrastive loss computation, the temperature parameter $\tau$ is set to 0.07. The total loss is calculated by the weighted combination of MSS loss, contrastive loss, and consistency loss, where the contrastive loss weight $\lambda_\mathrm{cont}$ and consistency loss weight $\lambda_\mathrm{consist}$ in~\Cref{eq:total_loss} are both set to 0.1 to maintain numerical balance between different loss components.

\subsection{Comparison with SOTA Methods}

We compare the proposed method with SOTA methods on the aforementioned four datasets under five different labeling ratios (5\%, 10\%, 20\%, 35\% and 50\%). The competitors include U-Net~\cite{unet_ronneberger_2015}, MT~\cite{meanteacher_tarvainen_2017}, UA-MT~\cite{uamt_yu_2019}, CCT~\cite{cct_ouali_2020}, MC-Net~\cite{mcnet_wu_2022}, SS-Net~\cite{ssnet_wu_2022}, BCP~\cite{bcp_bai_2023}, ABD~\cite{abd_chi_2024} and $\beta$-FFT~\cite{betafft_hu_2025}. The U-Net is also trained and tested in a fully supervised manner as a baseline.

Notably, the number of trainable parameters of CCT~\cite{cct_ouali_2020}, MC-Net~\cite{mcnet_wu_2022}, ABD~\cite{abd_chi_2024} and $\beta$-FFT~\cite{betafft_hu_2025} are significantly larger than the proposed method and other competitors, reaching 3.7M, 3.8M, 3.6M and 3.6M, respectively, while others adopt vanilla U-Net architecture with around 1.8M parameters. This difference may bring a potential performance bias between methods with different model complexities.

Quantitative performance comparisons between our proposed method Switch and previous SOTAs are reported in~\Cref{tab:results-1} (for LN-INT, LN-EXT, BUSI datasets) and~\Cref{tab:results-2} (for DDTI, TN3K, Prostate datasets).

\begin{table*}[htb]
	\caption{Experimental results of LN-INT, LN-EXT and BUSI datasets, averaged over three runs. `Params' denotes the number of trainable parameters, and `\#Label' refers to the labeled training ratio. The \textcolor{red}{best} and \textcolor{blue}{second best} results are highlighted, with $\uparrow$/$\downarrow$ indicating whether higher or lower values are better.}
	\centering
	\scriptsize
	\setlength{\tabcolsep}{4pt}
	\begin{tabular}{lc|c|cccc|cccc|cccc}
		\toprule
		\multirow{2}{*}{Method \tiny{(Venue)}} & \multirow{2}{*}{Params} & \multirow{2}{*}{\#Label } & \multicolumn{4}{c|}{LN-INT} & \multicolumn{4}{c|}{LN-EXT} & \multicolumn{4}{c}{BUSI}                                                                                                                                                                                    \\\cmidrule{4-15}
		                                       &                         &                           & Dice$\uparrow$              & IoU$\uparrow$               & HD95$\downarrow$         & ASD$\downarrow$  & Dice$\uparrow$    & IoU$\uparrow$     & HD95$\downarrow$  & ASD$\downarrow$   & Dice$\uparrow$    & IoU$\uparrow$     & HD95$\downarrow$  & ASD$\downarrow$   \\
		\midrule
		U-Net \tiny{(MICCAI'15)}               & 1.8 M                   & \multirow{10}{*}{5\%}     & 59.32                       & 49.55                       & 52.98                    & 17.82            & 48.92             & 39.26             & 62.76             & 21.64             & 40.28             & 30.28             & 70.27             & 27.77             \\
		MT \tiny{(NIPS'17)}                    & 1.8 M                   &                           & 57.45                       & 47.13                       & 56.84                    & 19.98            & 48.01             & 37.84             & 76.50             & 29.24             & 41.06             & 30.71             & 76.75             & 29.52             \\
		UA-MT \tiny{(MICCAI'19)}               & 1.8 M                   &                           & 55.42                       & 45.35                       & 58.75                    & 20.60            & 50.22             & 40.34             & 68.24             & 24.91             & 40.30             & 29.91             & 74.17             & 29.58             \\
		CCT \tiny{(CVPR'20)}                   & 3.7 M                   &                           & 58.42                       & 48.67                       & 50.37                    & 16.35            & 47.93             & 37.38             & 78.52             & 30.76             & 40.72             & 30.06             & 75.69             & 29.12             \\
		MC-Net \tiny{(MICCAI'21)}              & 3.8 M                   &                           & 61.01                       & 51.11                       & 51.70                    & 17.96            & 50.30             & 39.62             & 68.85             & 25.56             & 40.12             & 29.83             & 73.23             & 28.05             \\
		SS-Net \tiny{(MICCAI'22)}              & 1.8 M                   &                           & 53.84                       & 45.45                       & 48.45                    & 14.65            & 42.43             & 33.78             & 56.05             & 16.19             & 45.73             & 36.25             & 57.03             & 17.40             \\
		BCP \tiny{(CVPR'23)}                   & 1.8 M                   &                           & \color{blue}77.09           & 68.29                       & 34.56                    & 11.94            & 65.92             & 55.78             & 47.36             & 16.87             & 53.76             & 42.94             & 56.68             & 18.01             \\
		ABD \tiny{(CVPR'24)}                   & 3.6 M                   &                           & 76.62                       & \color{blue}68.55           & \color{blue}31.98        & \color{blue}9.63 & \color{blue}71.54 & \color{blue}62.75 & \color{blue}37.30 & \color{blue}11.17 & \color{blue}59.28 & \color{blue}49.87 & \color{blue}42.90 & \color{red}13.74  \\
		$\beta$-FFT \tiny{(CVPR'25)}           & 3.6 M                   &                           & 73.23                       & 63.94                       & 38.46                    & 11.75            & 62.33             & 52.60             & 46.85             & 12.40             & 54.64             & 44.99             & 44.61             & 15.66             \\
		\textbf{Switch \tiny{(Ours)}}          & 1.8 M                   &                           & \color{red}80.04            & \color{red}71.98            & \color{red}26.51         & \color{red}8.35  & \color{red}75.11  & \color{red}66.72  & \color{red}32.26  & \color{red}9.79   & \color{red}59.72  & \color{red}50.55  & \color{red}41.25  & \color{blue}13.96 \\
		\midrule
		U-Net \tiny{(MICCAI'15)}               & 1.8 M                   & \multirow{10}{*}{10\%}    & 70.17                       & 60.12                       & 50.17                    & 16.79            & 55.13             & 45.04             & 60.05             & 20.58             & 49.16             & 38.56             & 60.06             & 22.35             \\
		MT \tiny{(NIPS'17)}                    & 1.8 M                   &                           & 66.70                       & 57.11                       & 42.76                    & 15.13            & 54.56             & 43.98             & 56.89             & 19.83             & 43.78             & 33.88             & 62.21             & 22.45             \\
		UA-MT \tiny{(MICCAI'19)}               & 1.8 M                   &                           & 61.08                       & 52.20                       & 42.38                    & 13.83            & 50.64             & 40.98             & 61.36             & 21.83             & 46.19             & 35.12             & 67.15             & 25.73             \\
		CCT \tiny{(CVPR'20)}                   & 3.7 M                   &                           & 66.70                       & 57.19                       & 42.85                    & 14.13            & 55.55             & 44.46             & 65.29             & 24.42             & 49.25             & 38.46             & 58.29             & 22.19             \\
		MC-Net \tiny{(MICCAI'21)}              & 3.8 M                   &                           & 72.18                       & 61.98                       & 44.94                    & 15.30            & 56.89             & 46.20             & 63.94             & 23.21             & 45.44             & 36.15             & 59.07             & 21.85             \\
		SS-Net \tiny{(MICCAI'22)}              & 1.8 M                   &                           & 67.02                       & 57.46                       & 45.70                    & 14.69            & 52.13             & 41.87             & 64.71             & 20.47             & 49.35             & 38.76             & 64.45             & 24.74             \\
		BCP \tiny{(CVPR'23)}                   & 1.8 M                   &                           & 76.58                       & 67.83                       & 35.47                    & 11.79            & 68.63             & 58.33             & 47.35             & 16.60             & 57.81             & 48.09             & 45.99             & 17.25             \\
		ABD \tiny{(CVPR'24)}                   & 3.6 M                   &                           & \color{blue}81.26           & \color{blue}72.86           & 32.28                    & 10.71            & \color{blue}75.34 & \color{blue}66.26 & 36.96             & 10.99             & \color{blue}61.59 & \color{blue}52.74 & \color{red}37.75  & \color{red}13.01  \\
		$\beta$-FFT \tiny{(CVPR'25)}           & 3.6 M                   &                           & 78.27                       & 69.96                       & \color{blue}29.01        & \color{blue}8.37 & 73.30             & 64.39             & \color{blue}34.12 & \color{blue}10.08 & 58.88             & 50.01             & 41.01             & 15.72             \\
		\textbf{Switch \tiny{(Ours)}}          & 1.8 M                   &                           & \color{red}81.37            & \color{red}73.85            & \color{red}25.33         & \color{red}7.99  & \color{red}77.11  & \color{red}68.63  & \color{red}31.44  & \color{red}9.90   & \color{red}63.09  & \color{red}54.83  & \color{blue}37.81 & \color{blue}14.15 \\
		\midrule
		MT \tiny{(NIPS'17)}                    & 1.8 M                   &                           & 68.24                       & 59.88                       & 38.51                    & 13.36            & 66.34             & 56.64             & 51.96             & 17.55             & 56.48             & 46.03             & 66.82             & 28.22             \\
		U-Net \tiny{(MICCAI'15)}               & 1.8 M                   & \multirow{10}{*}{20\%}    & 73.14                       & 64.21                       & 42.79                    & 14.74            & 58.66             & 48.46             & 60.88             & 21.53             & 56.73             & 46.31             & 60.93             & 25.36             \\
		UA-MT \tiny{(MICCAI'19)}               & 1.8 M                   &                           & 71.35                       & 62.24                       & 43.89                    & 15.12            & 60.89             & 50.34             & 63.72             & 22.39             & 55.83             & 45.63             & 59.03             & 24.66             \\
		CCT \tiny{(CVPR'20)}                   & 3.7 M                   &                           & 71.11                       & 62.45                       & 36.11                    & 12.19            & 60.15             & 49.42             & 61.01             & 22.99             & 57.93             & 47.63             & 56.05             & 23.66             \\
		MC-Net \tiny{(MICCAI'21)}              & 3.8 M                   &                           & 73.44                       & 64.67                       & 37.21                    & 12.97            & 62.73             & 52.45             & 57.82             & 19.86             & 55.86             & 45.75             & 54.81             & 20.79             \\
		SS-Net \tiny{(MICCAI'22)}              & 1.8 M                   &                           & 72.75                       & 64.35                       & 42.98                    & 14.19            & 64.40             & 53.61             & 66.09             & 22.59             & 59.11             & 49.08             & 58.10             & 24.43             \\
		BCP \tiny{(CVPR'23)}                   & 1.8 M                   &                           & 78.21                       & 69.49                       & 34.08                    & 10.89            & 71.48             & 61.39             & 44.18             & 15.35             & 62.78             & 52.96             & 49.69             & 19.35             \\
		ABD \tiny{(CVPR'24)}                   & 3.6 M                   &                           & \color{blue}81.28           & \color{blue}73.84           & \color{blue}25.28        & \color{blue}7.50 & \color{blue}76.21 & \color{blue}67.90 & \color{blue}31.94 & \color{blue}9.92  & 66.80             & \color{blue}58.47 & \color{red}37.07  & \color{red}14.61  \\
		$\beta$-FFT \tiny{(CVPR'25)}           & 3.6 M                   &                           & 80.08                       & 71.35                       & 31.69                    & 9.63             & 73.79             & 64.99             & 35.99             & 11.34             & \color{blue}66.97 & 57.72             & 44.71             & 16.26             \\
		\textbf{Switch \tiny{(Ours)}}          & 1.8 M                   &                           & \color{red}82.98            & \color{red}75.86            & \color{red}20.41         & \color{red}6.75  & \color{red}79.10  & \color{red}70.95  & \color{red}29.08  & \color{red}9.09   & \color{red}67.53  & \color{red}58.61  & \color{blue}40.72 & \color{blue}16.08 \\
		\midrule
		U-Net \tiny{(MICCAI'15)}               & 1.8 M                   & \multirow{10}{*}{35\%}    & 76.57                       & 68.05                       & 35.55                    & 12.30            & 62.60             & 53.55             & 48.74             & 16.81             & 60.93             & 51.52             & 52.81             & 22.62             \\
		MT \tiny{(NIPS'17)}                    & 1.8 M                   &                           & 74.99                       & 66.30                       & 32.22                    & 10.84            & 60.02             & 50.58             & 48.07             & 17.01             & 62.67             & 53.76             & 53.24             & 19.66             \\
		UA-MT \tiny{(MICCAI'19)}               & 1.8 M                   &                           & 73.00                       & 64.54                       & 36.45                    & 11.71            & 57.39             & 47.90             & 52.62             & 19.66             & 62.40             & 52.93             & 48.77             & 18.84             \\
		CCT \tiny{(CVPR'20)}                   & 3.7 M                   &                           & 76.69                       & 67.70                       & 37.91                    & 12.85            & 61.56             & 52.05             & 50.40             & 17.86             & 62.31             & 53.26             & 49.30             & 21.59             \\
		MC-Net \tiny{(MICCAI'21)}              & 3.8 M                   &                           & 75.28                       & 68.08                       & 28.90                    & 9.01             & 61.98             & 53.17             & 46.44             & 15.76             & 61.37             & 52.14             & 45.64             & 16.78             \\
		SS-Net \tiny{(MICCAI'22)}              & 1.8 M                   &                           & 77.69                       & 70.19                       & 30.21                    & 9.54             & 61.05             & 51.78             & 66.23             & 20.89             & 62.66             & 53.93             & 45.20             & 17.60             \\
		BCP \tiny{(CVPR'23)}                   & 1.8 M                   &                           & 77.79                       & 69.53                       & 33.90                    & 11.15            & 69.34             & 59.12             & 45.42             & 16.36             & 64.03             & 54.11             & 49.07             & 18.12             \\
		ABD \tiny{(CVPR'24)}                   & 3.6 M                   &                           & \color{blue}82.26           & \color{blue}75.19           & \color{blue}22.64        & \color{blue}7.50 & \color{blue}77.04 & \color{blue}68.71 & \color{blue}33.21 & 11.61             & \color{blue}66.91 & \color{blue}58.61 & \color{red}37.92  & \color{red}13.76  \\
		$\beta$-FFT \tiny{(CVPR'25)}           & 3.6 M                   &                           & 80.42                       & 72.69                       & 27.52                    & 8.93             & 76.19             & 67.35             & 34.13             & \color{blue}10.79 & 66.74             & 58.12             & \color{blue}39.91 & \color{blue}16.71 \\
		\textbf{Switch \tiny{(Ours)}}          & 1.8 M                   &                           & \color{red}84.69            & \color{red}77.86            & \color{red}21.42         & \color{red}7.15  & \color{red}79.74  & \color{red}71.79  & \color{red}28.19  & \color{red}8.92   & \color{red}68.58  & \color{red}60.27  & 41.04             & 17.00             \\
		\midrule
		U-Net \tiny{(MICCAI'15)}               & 1.8 M                   & \multirow{10}{*}{50\%}    & 79.58                       & 71.89                       & 23.57                    & 7.74             & 65.18             & 56.21             & 43.35             & 14.72             & 62.80             & 52.98             & 52.15             & 20.94             \\
		MT \tiny{(NIPS'17)}                    & 1.8 M                   &                           & 78.84                       & 71.10                       & 28.26                    & 9.16             & 63.86             & 54.38             & 47.08             & 16.92             & 65.42             & 55.72             & 48.77             & 18.38             \\
		UA-MT \tiny{(MICCAI'19)}               & 1.8 M                   &                           & 78.69                       & 71.12                       & 26.31                    & 8.48             & 65.16             & 55.50             & 48.49             & 17.14             & 64.74             & 55.04             & 50.77             & 18.79             \\
		CCT \tiny{(CVPR'20)}                   & 3.7 M                   &                           & 77.32                       & 69.29                       & 27.05                    & 8.67             & 64.77             & 54.90             & 51.28             & 18.82             & 64.10             & 55.08             & 45.91             & 17.43             \\
		MC-Net \tiny{(MICCAI'21)}              & 3.8 M                   &                           & 79.12                       & 71.21                       & 25.43                    & 8.04             & 65.01             & 55.64             & 45.90             & 15.33             & 63.64             & 53.72             & 53.46             & 21.32             \\
		SS-Net \tiny{(MICCAI'22)}              & 1.8 M                   &                           & 80.03                       & 72.69                       & 24.43                    & 7.20             & 62.55             & 53.50             & 49.77             & 16.65             & 66.95             & 57.47             & 46.52             & 16.91             \\
		BCP \tiny{(CVPR'23)}                   & 1.8 M                   &                           & 80.35                       & 72.76                       & 23.40                    & \color{blue}7.09 & 71.94             & 62.02             & 41.22             & 14.06             & 67.14             & 57.23             & 44.49             & 16.47             \\
		ABD \tiny{(CVPR'24)}                   & 3.6 M                   &                           & \color{blue}83.37           & \color{blue}76.12           & \color{blue}22.88        & 7.29             & \color{blue}77.78 & \color{blue}69.53 & \color{blue}33.59 & 11.49             & \color{red}70.31  & 61.36             & 42.56             & 16.61             \\
		$\beta$-FFT \tiny{(CVPR'25)}           & 3.6 M                   &                           & 81.83                       & 74.18                       & 27.39                    & 8.83             & 74.62             & 66.33             & 35.54             & \color{blue}10.39 & 69.77             & \color{blue}61.50 & \color{red}35.91  & \color{red}13.50  \\
		\textbf{Switch \tiny{(Ours)}}          & 1.8 M                   &                           & \color{red}84.29            & \color{red}77.30            & \color{red}20.89         & \color{red}6.88  & \color{red}78.83  & \color{red}70.94  & \color{red}29.42  & \color{red}8.97   & \color{blue}70.24 & \color{red}61.57  & \color{blue}37.95 & \color{blue}15.61 \\
		\midrule
		U-Net \tiny{(MICCAI'15)}               & 1.8 M                   & 100\%                     & 83.49                       & 76.61                       & 22.39                    & 7.06             & 70.00             & 60.65             & 42.21             & 14.14             & 68.85             & 59.55             & 42.65             & 15.28             \\
		\bottomrule
	\end{tabular}
	\label{tab:results-1}
\end{table*}

\begin{table*}[htb]
	\caption{Experimental results on the DDTI, TN3K, and Prostate datasets, averaged over three runs. `Params' denotes the number of trainable parameters, and `\#Label' refers to the labeled training ratio. The \textcolor{red}{best} and \textcolor{blue}{second best} results are highlighted, with $\uparrow$/$\downarrow$ indicating whether higher or lower values are better.}
	\centering
	\scriptsize
	\setlength{\tabcolsep}{4pt}
	\begin{tabular}{lc|c|cccc|cccc|cccc}
		\toprule
		\multirow{2}{*}{Method \tiny{(Venue)}} & \multirow{2}{*}{Params} & \multirow{2}{*}{\#Label } & \multicolumn{4}{c|}{DDTI} & \multicolumn{4}{c|}{TN3K} & \multicolumn{4}{c}{Prostate}                                                                                                                                                                                    \\\cmidrule{4-15}
		                                       &                         &                           & Dice$\uparrow$            & IoU$\uparrow$             & HD95$\downarrow$             & ASD$\downarrow$   & Dice$\uparrow$    & IoU$\uparrow$     & HD95$\downarrow$  & ASD$\downarrow$   & Dice$\uparrow$    & IoU$\uparrow$     & HD95$\downarrow$  & ASD$\downarrow$  \\
		\midrule
		U-Net \tiny{(MICCAI'15)}               & 1.8 M                   & \multirow{10}{*}{5\%}     & 76.31                     & 62.51                     & 39.49                        & 15.64             & 54.99             & 43.77             & 52.31             & 20.70             & 79.84             & 73.40             & 36.38             & 11.28            \\
		MT \tiny{(NIPS'17)}                    & 1.8 M                   &                           & 75.74                     & 61.83                     & 41.05                        & 15.89             & 55.50             & 44.03             & 53.10             & 20.30             & 78.99             & 72.06             & 38.48             & 12.05            \\
		UA-MT \tiny{(MICCAI'19)}               & 1.8 M                   &                           & 76.17                     & 62.39                     & 42.08                        & 16.23             & 56.86             & 45.16             & 52.82             & 20.24             & 78.83             & 71.61             & 38.45             & 12.16            \\
		CCT \tiny{(CVPR'20)}                   & 3.7 M                   &                           & 77.50                     & 64.25                     & 45.57                        & 18.79             & 59.79             & 48.36             & 47.54             & 18.14             & 79.60             & 72.73             & 33.78             & 10.25            \\
		MC-Net \tiny{(MICCAI'21)}              & 3.8 M                   &                           & 80.88                     & 68.49                     & 32.89                        & 13.16             & 58.18             & 46.78             & 48.81             & 18.34             & 79.43             & 72.94             & 28.10             & 8.29             \\
		SS-Net \tiny{(MICCAI'22)}              & 1.8 M                   &                           & 75.99                     & 62.15                     & 39.59                        & 15.13             & 57.66             & 45.94             & 51.57             & 19.59             & 79.29             & 72.31             & 36.33             & 11.13            \\
		BCP \tiny{(CVPR'23)}                   & 1.8 M                   &                           & 82.07                     & 70.30                     & 31.52                        & 12.62             & 66.98             & 56.13             & 36.82             & 13.20             & \color{blue}83.33 & \color{blue}78.66 & 20.76             & 6.48             \\
		ABD \tiny{(CVPR'24)}                   & 3.6 M                   &                           & \color{blue}84.82         & \color{blue}74.47         & \color{red}27.41             & \color{blue}10.62 & \color{red}71.74  & \color{red}61.46  & \color{blue}32.54 & \color{blue}11.63 & 83.12             & 78.51             & \color{blue}20.63 & \color{blue}6.30 \\
		$\beta$-FFT \tiny{(CVPR'25)}           & 3.6 M                   &                           & 83.76                     & 72.72                     & 28.32                        & 10.76             & 67.50             & 57.00             & 36.86             & 13.95             & 81.49             & 76.22             & 22.39             & 6.53             \\
		\textbf{Switch \tiny{(Ours)}}          & 1.8 M                   &                           & \color{red}85.52          & \color{red}75.46          & \color{blue}28.08            & \color{red}10.18  & \color{blue}71.72 & \color{blue}61.40 & \color{red}28.90  & \color{red}9.51   & \color{red}83.48  & \color{red}79.11  & \color{red}17.29  & \color{red}5.37  \\
		\midrule
		U-Net \tiny{(MICCAI'15)}               & 1.8 M                   & \multirow{10}{*}{10\%}    & 81.39                     & 69.36                     & 37.17                        & 13.75             & 62.07             & 51.06             & 44.09             & 16.69             & 81.14             & 75.33             & 30.25             & 9.32             \\
		MT \tiny{(NIPS'17)}                    & 1.8 M                   &                           & 79.81                     & 67.22                     & 40.03                        & 14.85             & 63.32             & 52.17             & 44.71             & 16.40             & 80.39             & 74.19             & 31.33             & 9.92             \\
		UA-MT \tiny{(MICCAI'19)}               & 1.8 M                   &                           & 79.88                     & 67.19                     & 40.33                        & 14.90             & 62.75             & 51.60             & 44.66             & 16.90             & 80.06             & 73.67             & 34.17             & 10.88            \\
		CCT \tiny{(CVPR'20)}                   & 3.7 M                   &                           & 78.31                     & 65.52                     & 45.41                        & 17.97             & 66.27             & 54.91             & 41.93             & 15.20             & 81.12             & 75.24             & 27.12             & 8.12             \\
		MC-Net \tiny{(MICCAI'21)}              & 3.8 M                   &                           & 82.21                     & 70.58                     & 34.78                        & 12.95             & 63.80             & 53.05             & 37.60             & 13.95             & 81.10             & 75.60             & 23.40             & 7.02             \\
		SS-Net \tiny{(MICCAI'22)}              & 1.8 M                   &                           & 81.89                     & 69.92                     & 35.09                        & 13.16             & 65.84             & 54.69             & 42.93             & 15.59             & 80.13             & 73.73             & 31.50             & 9.50             \\
		BCP \tiny{(CVPR'23)}                   & 1.8 M                   &                           & 82.62                     & 71.22                     & 32.82                        & 12.78             & 67.93             & 57.38             & 33.20             & 11.44             & \color{blue}83.57 & \color{blue}79.03 & 19.25             & 5.79             \\
		ABD \tiny{(CVPR'24)}                   & 3.6 M                   &                           & 84.90                     & 74.63                     & 30.64                        & 11.10             & \color{red}74.11  & \color{red}64.36  & \color{blue}29.23 & \color{blue}10.20 & 82.88             & 78.44             & \color{blue}19.22 & 5.88             \\
		$\beta$-FFT \tiny{(CVPR'25)}           & 3.6 M                   &                           & \color{blue}85.23         & \color{blue}75.06         & \color{red}28.17             & \color{red}10.25  & 70.19             & 60.11             & 30.14             & 10.52             & 81.78             & 77.01             & 20.06             & \color{blue}5.75 \\
		\textbf{Switch \tiny{(Ours)}}          & 1.8 M                   &                           & \color{red}85.39          & \color{red}75.49          & \color{blue}29.45            & \color{blue}10.67 & \color{blue}72.98 & \color{blue}63.17 & \color{red}26.51  & \color{red}8.63   & \color{red}83.63  & \color{red}79.41  & \color{red}17.11  & \color{red}5.23  \\
		\midrule
		U-Net \tiny{(MICCAI'15)}               & 1.8 M                   & \multirow{10}{*}{20\%}    & 84.47                     & 73.69                     & 30.00                        & 11.01             & 66.61             & 55.70             & 40.74             & 14.86             & 82.18             & 76.99             & 23.70             & 7.25             \\
		MT \tiny{(NIPS'17)}                    & 1.8 M                   &                           & 84.18                     & 73.28                     & 32.28                        & 11.56             & 67.65             & 56.82             & 36.44             & 13.24             & 80.48             & 74.37             & 28.15             & 8.79             \\
		UA-MT \tiny{(MICCAI'19)}               & 1.8 M                   &                           & 83.93                     & 72.87                     & 33.55                        & 11.94             & 66.89             & 56.03             & 38.88             & 14.34             & 81.26             & 75.51             & 25.44             & 7.91             \\
		CCT \tiny{(CVPR'20)}                   & 3.7 M                   &                           & 83.49                     & 72.57                     & 36.34                        & 13.04             & 67.56             & 56.73             & 36.14             & 12.89             & 82.10             & 76.67             & 24.89             & 7.26             \\
		MC-Net \tiny{(MICCAI'21)}              & 3.8 M                   &                           & 84.92                     & 74.53                     & 28.49                        & 10.56             & 68.69             & 58.13             & 33.47             & 11.34             & 80.88             & 75.25             & 22.62             & 6.86             \\
		SS-Net \tiny{(MICCAI'22)}              & 1.8 M                   &                           & 84.15                     & 73.11                     & 31.33                        & 11.48             & 68.18             & 57.64             & 36.20             & 12.96             & 80.65             & 74.62             & 28.17             & 8.62             \\
		BCP \tiny{(CVPR'23)}                   & 1.8 M                   &                           & 85.50                     & 75.34                     & 27.14                        & 10.02             & 70.70             & 60.35             & 30.75             & 10.49             & \color{blue}83.96 & 79.55             & 18.26             & 5.41             \\
		ABD \tiny{(CVPR'24)}                   & 3.6 M                   &                           & 87.76                     & 78.77                     & 24.07                        & 8.45              & \color{blue}75.08 & \color{blue}65.43 & \color{blue}26.26 & \color{blue}8.92  & 83.78             & \color{blue}79.73 & \color{red}15.80  & \color{red}4.80  \\
		$\beta$-FFT \tiny{(CVPR'25)}           & 3.6 M                   &                           & \color{red}88.28          & \color{red}79.45          & \color{red}23.04             & \color{red}7.76   & 71.39             & 61.67             & 26.59             & 9.07              & 81.25             & 75.87             & 26.38             & 7.58             \\
		\textbf{Switch \tiny{(Ours)}}          & 1.8 M                   &                           & \color{blue}87.78         & \color{blue}78.97         & \color{blue}23.93            & \color{blue}8.44  & \color{red}75.10  & \color{red}65.46  & \color{red}24.20  & \color{red}7.78   & \color{red}84.06  & \color{red}80.00  & \color{blue}16.02 & \color{blue}4.99 \\
		\midrule
		U-Net \tiny{(MICCAI'15)}               & 1.8 M                   & \multirow{10}{*}{35\%}    & 85.95                     & 75.98                     & 27.93                        & 9.75              & 70.13             & 59.76             & 34.79             & 12.21             & 82.97             & 78.07             & 21.69             & 6.69             \\
		MT \tiny{(NIPS'17)}                    & 1.8 M                   &                           & 86.08                     & 76.07                     & 29.05                        & 10.05             & 70.72             & 60.11             & 36.56             & 13.07             & 81.46             & 75.79             & 24.52             & 7.38             \\
		UA-MT \tiny{(MICCAI'19)}               & 1.8 M                   &                           & 85.61                     & 75.34                     & 30.48                        & 10.51             & 71.06             & 60.47             & 34.96             & 12.14             & 81.79             & 76.18             & 24.88             & 7.42             \\
		CCT \tiny{(CVPR'20)}                   & 3.7 M                   &                           & 84.84                     & 74.43                     & 34.99                        & 12.06             & 69.72             & 59.00             & 32.91             & 11.64             & 82.79             & 77.81             & 20.07             & 5.97             \\
		MC-Net \tiny{(MICCAI'21)}              & 3.8 M                   &                           & 86.82                     & 77.29                     & 27.26                        & 9.13              & 70.27             & 60.25             & 32.37             & 11.04             & 82.10             & 76.84             & 20.49             & 6.18             \\
		SS-Net \tiny{(MICCAI'22)}              & 1.8 M                   &                           & 86.06                     & 76.08                     & 27.76                        & 9.70              & 71.76             & 61.30             & 31.91             & 11.09             & 82.37             & 76.99             & 23.90             & 7.08             \\
		BCP \tiny{(CVPR'23)}                   & 1.8 M                   &                           & 86.74                     & 77.22                     & 25.67                        & 8.96              & 73.29             & 63.12             & 27.98             & 8.99              & \color{blue}84.15 & \color{blue}80.00 & \color{blue}16.87 & \color{blue}5.06 \\
		ABD \tiny{(CVPR'24)}                   & 3.6 M                   &                           & \color{red}88.97          & \color{red}80.64          & \color{red}21.63             & \color{red}7.22   & \color{red}76.99  & \color{red}67.43  & \color{blue}24.42 & \color{blue}7.97  & 83.66             & 79.54             & 17.72             & 5.49             \\
		$\beta$-FFT \tiny{(CVPR'25)}           & 3.6 M                   &                           & 88.28                     & 79.45                     & 23.04                        & \color{blue}7.76  & 72.56             & 62.69             & 27.26             & 9.91              & 81.30             & 76.46             & 20.61             & 6.05             \\
		\textbf{Switch \tiny{(Ours)}}          & 1.8 M                   &                           & \color{blue}88.70         & \color{blue}80.30         & \color{blue}22.64            & 7.80              & \color{blue}75.89 & \color{blue}66.55 & \color{red}22.51  & \color{red}6.91   & \color{red}84.25  & \color{red}80.29  & \color{red}15.17  & \color{red}4.76  \\
		\midrule
		U-Net \tiny{(MICCAI'15)}               & 1.8 M                   & \multirow{10}{*}{50\%}    & 87.02                     & 77.60                     & 25.36                        & 8.89              & 71.72             & 61.67             & 31.40             & 11.07             & 82.90             & 78.08             & 19.91             & 6.00             \\
		MT \tiny{(NIPS'17)}                    & 1.8 M                   &                           & 86.55                     & 76.86                     & 27.42                        & 9.28              & 72.39             & 62.22             & 31.62             & 10.94             & 81.74             & 76.27             & 21.83             & 6.71             \\
		UA-MT \tiny{(MICCAI'19)}               & 1.8 M                   &                           & 86.84                     & 77.28                     & 27.90                        & 9.51              & 72.46             & 62.17             & 32.47             & 11.39             & 81.77             & 76.41             & 21.18             & 6.55             \\
		CCT \tiny{(CVPR'20)}                   & 3.7 M                   &                           & 85.94                     & 76.08                     & 30.28                        & 10.65             & 72.63             & 62.50             & 29.76             & 10.51             & 82.73             & 77.69             & 19.57             & 5.74             \\
		MC-Net \tiny{(MICCAI'21)}              & 3.8 M                   &                           & 87.31                     & 78.05                     & 24.94                        & 8.55              & 73.03             & 63.05             & 27.49             & 8.96              & 82.41             & 77.37             & 19.49             & 5.62             \\
		SS-Net \tiny{(MICCAI'22)}              & 1.8 M                   &                           & 86.65                     & 76.98                     & 26.70                        & 9.18              & 73.53             & 63.32             & 30.75             & 10.72             & 82.06             & 76.63             & 24.10             & 7.05             \\
		BCP \tiny{(CVPR'23)}                   & 1.8 M                   &                           & 87.23                     & 78.01                     & 26.12                        & 8.88              & 74.59             & 64.46             & 28.20             & 9.26              & \color{blue}83.97 & \color{blue}79.75 & \color{blue}16.62 & \color{blue}4.95 \\
		ABD \tiny{(CVPR'24)}                   & 3.6 M                   &                           & \color{blue}89.20         & \color{blue}80.97         & \color{blue}20.59            & \color{red}7.12   & \color{blue}76.15 & \color{blue}66.55 & \color{blue}23.64 & \color{blue}7.45  & 82.97             & 78.22             & 20.07             & 5.93             \\
		$\beta$-FFT \tiny{(CVPR'25)}           & 3.6 M                   &                           & 88.33                     & 79.59                     & 22.73                        & 7.76              & 73.92             & 64.20             & 25.31             & 8.70              & 82.48             & 78.11             & 16.91             & 5.04             \\
		\textbf{Switch \tiny{(Ours)}}          & 1.8 M                   &                           & \color{red}89.26          & \color{red}81.12          & \color{red}20.32             & \color{blue}7.18  & \color{red}76.24  & \color{red}66.82  & \color{red}22.12  & \color{red}6.96   & \color{red}84.21  & \color{red}80.30  & \color{red}14.68  & \color{red}4.61  \\
		\midrule
		U-Net \tiny{(MICCAI'15)}               & 1.8 M                   & 100\%                     & 87.95                     & 78.99                     & 24.54                        & 8.34              & 74.77             & 64.86             & 28.32             & 9.37              & 82.82             & 78.19             & 18.59             & 5.60             \\
		\bottomrule
	\end{tabular}
	\label{tab:results-2}
\end{table*}

\begin{figure*}[htb]
	\centering
	\includegraphics[width=1\linewidth]{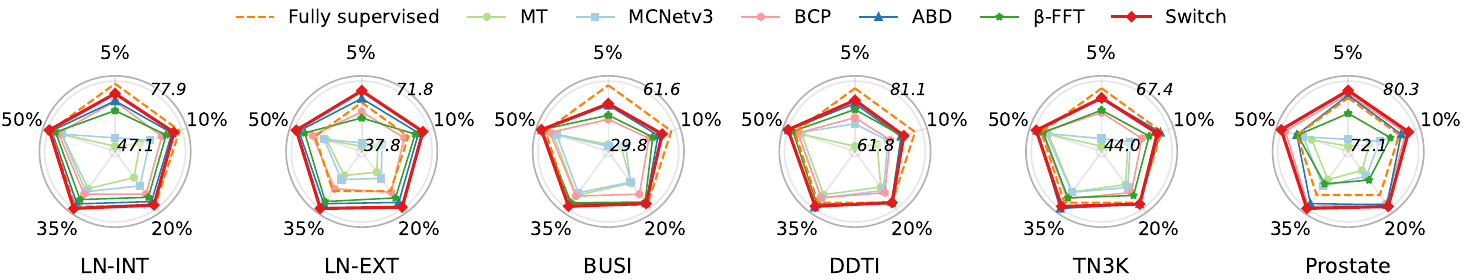}
	\caption{A comparison of IoU values of the proposed method and SOTAs on six datasets with five data labeling ratios. The maximum and minimum IoU values obtained by these methods for each dataset are marked in italics. All labeling ratios are displayed in the five corresponding corners of the radar charts.}
	\label{fig:radar}
\end{figure*}

\textbf{Overall Performance Analysis}. Across six diverse US datasets, our proposed Switch method demonstrates consistent superiority in segmentation accuracy. As illustrated in the comprehensive IoU radar chart analysis (\Cref{fig:radar}), Switch consistently maintains competitive or superior performance across all labeling ratios on all datasets. The visualization reveals three key patterns: (1) substantial improvements in low-data regimes, (2) consistent parameter efficiency advantages, and (3) strong cross-domain generalization capabilities demonstrated across diverse anatomical targets (lymph nodes, breast lesions, thyroid nodules, and prostate regions).

\textbf{Low-Data Regime Excellence}. The most significant advantages emerge under limited labeling scenarios (5-10\%). At 5\% labeling ratio group, Switch achieves remarkable performance: LN-INT (80.04\% Dice, 71.98\% IoU), DDTI (85.52\% Dice, 75.46\% IoU), and Prostate (83.48\% Dice, 79.11\% IoU). Compared to the strongest competitors, Switch shows consistent improvements ranging from +0.15\% (Prostate) to +3.57\% (LN-EXT) Dice across datasets. Boundary precision metrics further validate our superiority, with Switch achieving the lowest HD95 values on most datasets, including LN-INT (26.51 vs. ABD 31.98) and Prostate (17.29 vs. BCP 20.76).

\textbf{Cross-Domain Generalization}. The external validation on LN-EXT provides crucial evidence of generalization capability across hospital centers. Switch maintains consistent Dice advantages: +3.57\% (5\%), +1.77\% (10\%), +2.89\% (20\%), +2.70\% (35\%), and +1.05\% (50\%) compared to the second-best method ABD~\cite{abd_chi_2024}. The boundary metrics show similar robustness with HD95 improvements consistently exceeding 3.0 across all ratios.

\textbf{Parameter Efficiency Validation}. Switch consistently outperforms parameter-equivalent baselines (1.8M parameters) across all scenarios. Against BCP~\cite{bcp_bai_2023}, notable improvements include: LN-EXT (+9.19\% Dice), BUSI (+5.96\% Dice), and DDTI (+3.45\% Dice) at 5\% labeling. When compared to 3.6M+ parameter methods, Switch often matches or exceeds performance while requiring only half the computational resources, exemplified by the 8.04\% IoU improvement over $\beta$-FFT~\cite{betafft_hu_2025} on LN-INT.

\begin{figure*}[htb]
	\centering
	\includegraphics[width=0.9\linewidth]{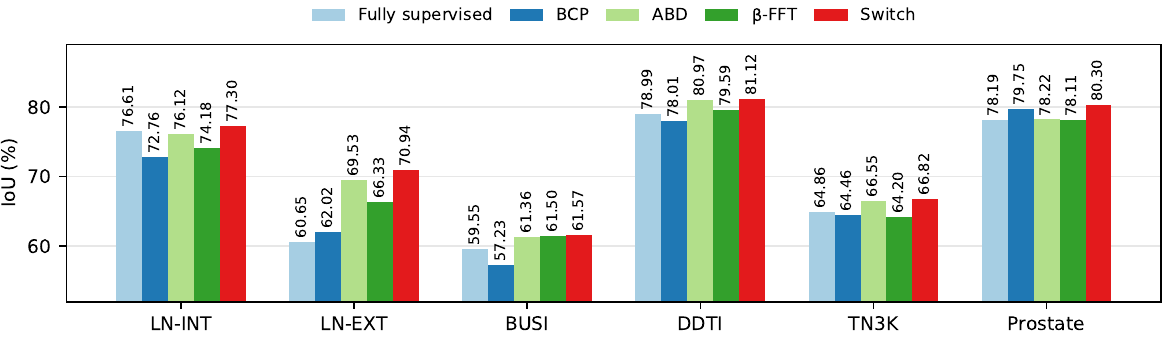}
	\caption{A comparison of IoU values of the proposed method and SOTAs on six datasets with 50\% labeled data. IoU values of all methods are displayed above the corresponding bar charts.}
	\label{fig:bar}
\end{figure*}

\textbf{High-Data Regime Validation}. The comprehensive IoU comparison at 50\% labeling (shown in~\Cref{fig:bar}) demonstrates consistent superiority even in data-rich scenarios. Switch achieves the highest IoU across all datasets: 77.30\% (LN-INT), 70.94\% (LN-EXT), 61.57\% (BUSI), 81.12\% (DDTI), 66.82\% (TN3K), and 80.30\% (Prostate). Remarkably, our semi-supervised approach with 50\% labeled data not only consistently outperforms the strongest competitors across all datasets, with advantages ranging from 0.07\% (BUSI, $\beta$-FFT) to 1.41\% (LN-EXT, ABD), but also exceeds fully supervised training on LN-INT (+0.69\%) and LN-EXT (+10.29\%), demonstrating the effectiveness of our unlabeled data utilization strategy even in data-rich scenarios.

\textbf{Summary}. The quantitative evaluation across six US datasets establishes three key contributions: (1) \textit{Superior low-data performance}: Switch consistently achieves optimal results at 5-10\% labeling, with 20\% labeled performance exceeding fully supervised baselines on multiple datasets; (2) \textit{Parameter efficiency}: our 1.8M-parameter model outperforms equivalent-complexity methods by 2-9\% Dice while matching 3.6M-parameter methods; (3) \textit{Cross-domain robustness}: maintaining consistent advantages on external validation across all labeling ratios. These results validate the effectiveness of our approach for resource-constrained medical imaging applications.

\subsection{Visualization of Segmentation Results}

\begin{figure*}[htb]
	\centering
	\includegraphics[width=1\linewidth]{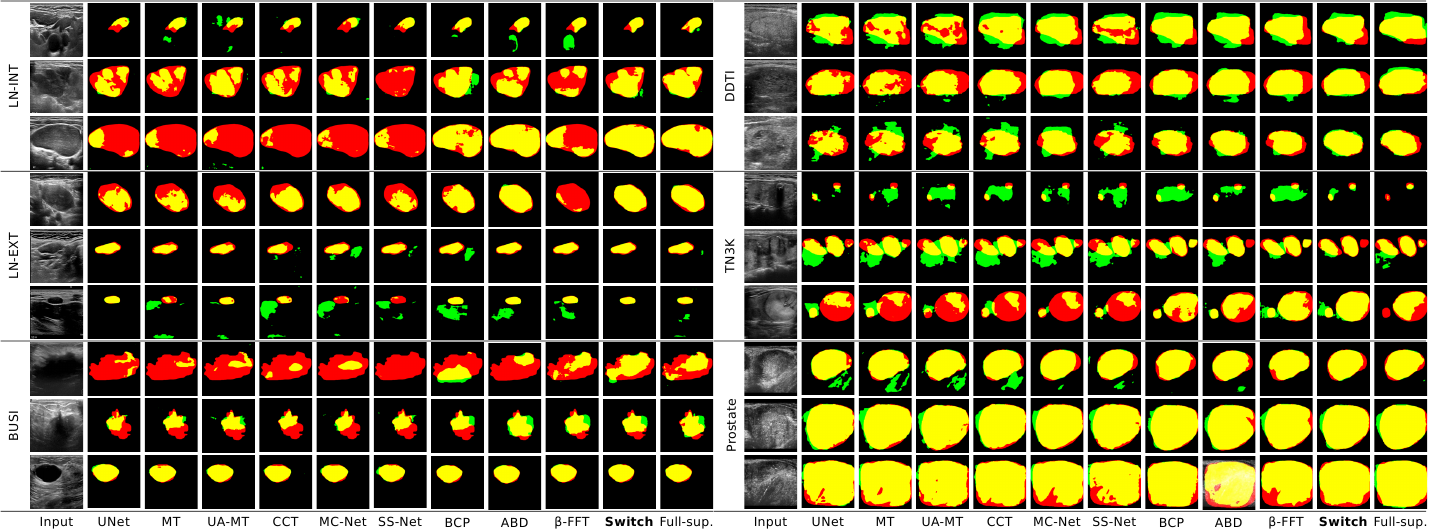}
	\caption{Visualization of the proposed method and SOTAs on LN-INT, LN-EXT, BUSI~\cite{busi_al_2020}, DDTI~\cite{ddti_pedraza_2015}, TN3K~\cite{tn3k_gong_2023} and Prostate~\cite{microsegnet_jiang_2023} datasets with 5\% labeled data. The first column shows the original images and Full-sup. refers to the results of U-Net trained in a fully supervised manner (\ie, 100\% labeled). Regions in red, green and yellow indicate the ground truth, false positive and true positive, respectively.}
	\label{fig:results}
\end{figure*}

\Cref{fig:results} depict the visualization of the segmentation results of the proposed method and the SOTA techniques on the LN-INT, LN-EXT, BUSI~\cite{busi_al_2020}, DDTI~\cite{ddti_pedraza_2015}, TN3K~\cite{tn3k_gong_2023} and Prostate~\cite{microsegnet_jiang_2023} datasets with only 5\% labeled data available.

The qualitative results reveal three distinct advantage patterns of our Switch method across challenging US scenarios:

\textbf{Low-Contrast Boundary Recovery}. In challenging cases with weak tissue contrast (e.g., BUSI breast lesions and TN3K thyroid nodules), competitors frequently produce fragmented or incomplete segmentations. Our method demonstrates superior boundary completeness, achieving near-complete lesion coverage that closely approximates the ground truth annotations. This improvement is particularly evident in the homogeneous tissue regions where traditional methods struggle to distinguish subtle intensity variations.

\textbf{Artifact Robustness}. US imaging artifacts, including shadowing, speckle noise, and irregular anatomical boundaries, pose significant challenges for automated segmentation. The visualization demonstrates that Switch maintains segmentation consistency across these problematic regions, with notably reduced false positive areas compared to existing methods. This robustness stems from the ability of FDS to preserve structural information while enhancing feature discriminability.

\textbf{Semi-Supervised vs. Fully-Supervised Performance}. Remarkably, our semi-supervised approach with only 5\% labeled data produces segmentation quality comparable to, and in some cases (LN-INT, LN-EXT) superior to, fully supervised U-Net training. This achievement validates the effectiveness of our unlabeled data utilization strategy, demonstrating that proper SSL can overcome the traditional performance gap between limited and full supervision scenarios.

\section{Discussion}\label{sec:discussion}

\subsection{Embedding Space Analysis}

\begin{figure}[htb]
	\centering
	\includegraphics[width=1\linewidth]{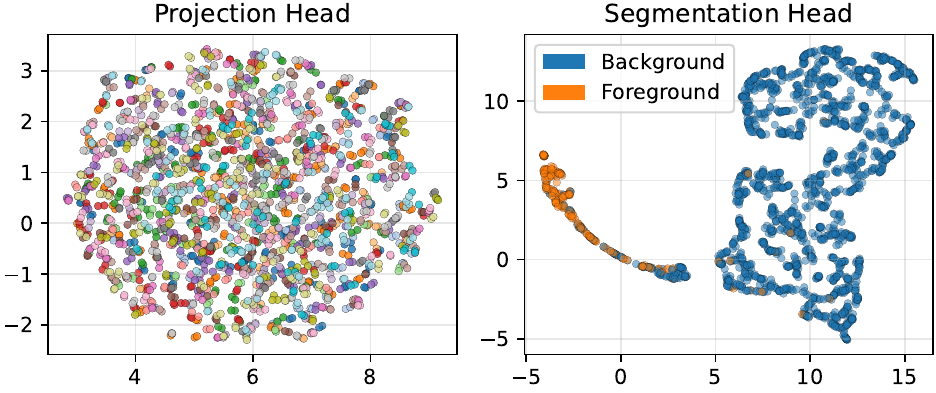}
	\caption{UMAP visualization of embedding spaces for projection head and segmentation head using 20 LN-INT test samples. In the projection head, dots of the same color represent feature pairs from original and FDS-reconstructed images, demonstrating their proximity after CL. In the segmentation embedding space, yellow and blue dots denote foreground and background features, respectively.}
	\label{fig:umap}
\end{figure}

To validate the effectiveness of different network components and address reviewer concerns about embedding space characteristics, we provide UMAP visualizations of both projection head and segmentation head features (\Cref{fig:umap}).

\textbf{Projection Head Analysis}. The projection head embedding reveals the effectiveness of our CL approach. Identical-colored pairs clustering together represent corresponding spatial positions from original and FDS-reconstructed images, validating that $\mathcal{L}_\mathrm{cont}$ successfully learns augmentation-invariant representations. Different colors maintain appropriate distances for effective instance discrimination. This embedding prioritizes spatial consistency over semantic classification, aligning with the instance-level objectives of CL.

\textbf{Segmentation Head Analysis}. The segmentation head demonstrates fundamentally different organization driven by supervised classification objectives. Through combined loss functions ($\mathcal{L}_\mathrm{dice}$ and $\mathcal{L}_\mathrm{ce}$), it learns to maximize inter-class separability. The visualization reveals two distinct clusters: dominant background features (blue) and compact foreground lesion features (yellow), reflecting the class imbalance in medical US imaging. This semantic clustering confirms successful discriminative feature learning for accurate pixel-level classification.

\subsection{Ablation Studies}

To investigate the individual contributions and optimal configurations of the proposed components, we conduct comprehensive ablation studies on the LN dataset with 5\% labeled data. The analysis covers module effectiveness validation, MSS patch configuration, and augmentation strategies.

\begin{figure}[htb]
	\centering
	\includegraphics[width=1\linewidth]{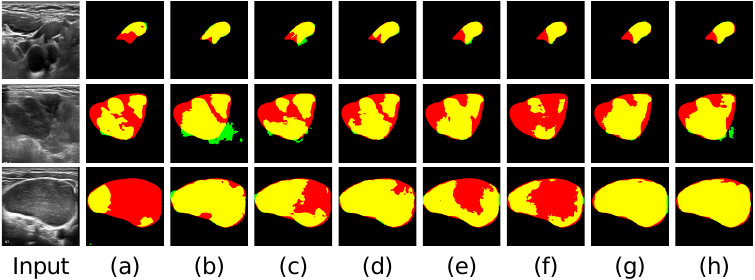}
	\caption{Visual comparison of ablation studies on LN-INT dataset with 5\% labeled data. The first column shows the original images and (a)-(h) refer to the experiments in~\Cref{tab:abla_modules}. Regions in red, green and yellow indicate the ground truth, false positive and true positive, respectively.}
	\label{fig:abla}
\end{figure}

\textbf{Module Effectiveness Validation}. \Cref{tab:abla_modules} provides comprehensive validation of the contribution from each proposed component. The baseline (a) achieves only 59.52\% Dice, while MSS alone (b) dramatically improves performance to 75.15\% Dice (+15.63\%), demonstrating its fundamental importance. Adding FDS with CL (c) slightly reduces performance (74.17\% Dice), but consistency regularization (d) recovers it to 75.65\% Dice. The inclusion of augmentation strategies (e) further boosts performance to 76.79\% Dice. The complete framework (h) achieves 80.04\% Dice, representing a remarkable +20.52\% improvement over the baseline. Notably, boundary precision metrics show consistent improvements, with HD95 decreasing from 55.57 (baseline) to 26.51 (full model), confirming that each module contributes to both overlap accuracy and boundary quality. The qualitative results in~\Cref{fig:abla} visually demonstrate these progressive improvements, where the complete framework (h) produces segmentations with significantly reduced false positives and improved boundary adherence compared to individual modules.

\begin{table}[htb]
	\caption{Ablation study of different modules on the LN dataset with 5\% labeled data. Augs., Cont. and Consist. refer to augmentation, contrastive and consistency, respectively.}
	\centering
	\footnotesize
	\setlength{\tabcolsep}{3pt}
	\begin{tabular}{c|ccc|cc|cccc}
		\toprule
		Exp. & MSS    & FDS    & Augs.  & Cont.  & Consist. & Dice$\uparrow$   & IoU$\uparrow$    & HD95$\downarrow$ & ASD$\downarrow$ \\
		\midrule
		(a)  &        &        &        &        &          & 59.52            & 49.86            & 55.57            & 18.82           \\
		(b)  & \cmark &        &        &        &          & 75.15            & 66.47            & 35.98            & 11.99           \\
		(c)  & \cmark & \cmark &        & \cmark &          & 74.17            & 65.26            & 36.54            & 11.64           \\
		(d)  & \cmark & \cmark &        &        & \cmark   & 75.65            & 66.91            & 35.67            & 11.80           \\
		(e)  & \cmark &        & \cmark &        &          & 76.79            & 68.74            & 28.42            & 9.55            \\
		\midrule
		(f)  & \cmark & \cmark & \cmark & \cmark &          & 78.24            & 70.00            & 29.52            & 10.30           \\
		(g)  & \cmark & \cmark & \cmark &        & \cmark   & 77.85            & 69.70            & 30.40            & 10.04           \\
		(h)  & \cmark & \cmark & \cmark & \cmark & \cmark   & \color{red}80.04 & \color{red}71.98 & \color{red}26.51 & \color{red}8.35 \\
		\bottomrule
	\end{tabular}
	\label{tab:abla_modules}
\end{table}

\textbf{Multiscale Switch Configuration}. We examine the optimal configuration of coarse and fine patches in MSS on segmentation performance. As shown in~\Cref{tab:abla_views}, the best performance is achieved with $p=2$ coarse patches and $q=2$ fine patches, reaching 80.04\% Dice and 71.98\% IoU. This configuration covers 17/32 (53\%) of the image area, providing balanced information exchange between labeled and unlabeled data. Increasing the number of patches beyond this optimal point leads to performance degradation, with HD95 increasing from 26.51 to 33.74 when using $p=3, q=4$. This suggests that excessive patch coverage may introduce noise and degrade boundary precision.

\begin{table}[htb]
	\caption{Ablation study of the amount of coarse and fine patches on the LN dataset with 5\% labeled data. The Max. area column indicates the maximum ratio of the total image area occupied by all patches without overlapping.}
	\centering
	\footnotesize
	\setlength{\tabcolsep}{3pt}
	\begin{tabular}{cc|cccc|c}
		\toprule
		\makecell{$p$\\coarse patches} & \makecell{$q$\\fine patches} & Dice$\uparrow$   & IoU$\uparrow$    & HD95$\downarrow$ & ASD$\downarrow$ & Max. area \\
		\midrule
		1                              & 2                            & 78.32            & 70.24            & 27.86            & 8.79            & 9/32      \\
		1                              & 4                            & 78.32            & 70.14            & 29.10            & 9.28            & 10/32     \\
		2                              & 2                            & \color{red}80.04 & \color{red}71.98 & \color{red}26.51 & \color{red}8.35 & 17/32     \\
		2                              & 4                            & 79.70            & 71.20            & 29.87            & 9.52            & 18/32     \\
		3                              & 2                            & 79.20            & 71.13            & 29.96            & 9.34            & 25/32     \\
		3                              & 4                            & 78.76            & 70.40            & 33.74            & 11.00           & 26/32     \\
		\bottomrule
	\end{tabular}
	\label{tab:abla_views}
\end{table}



\textbf{Augmentation Strategy Analysis}. \Cref{tab:abla_augs} demonstrates the critical role of augmentation strategies in our framework. Weak augmentation alone (random flip, resize, crop) provides substantial improvements of +1.92\% Dice and +1.81\% IoU over the baseline. Interestingly, strong augmentation alone proves detrimental (-0.24\% Dice, -1.21\% IoU), likely due to excessive distortion that conflicts with the CL objectives. However, the combination of weak and strong augmentations achieves optimal performance (80.04\% Dice, 71.98\% IoU), suggesting that weak augmentations provide necessary geometric variations while strong augmentations contribute complementary feature diversity when properly balanced.

\begin{table}[htb]
	\caption{Ablation study of different augmentations on the LN dataset with 5\% labeled data.}
	\centering
	\footnotesize
	\setlength{\tabcolsep}{4pt}
	\begin{tabular}{cc|cccc}
		\toprule
		Weak   & Strong & Dice$\uparrow$   & IoU$\uparrow$    & HD95$\downarrow$ & ASD$\downarrow$ \\
		\midrule
		       &        & 77.38            & 69.51            & 30.32            & 9.42            \\
		\cmark &        & 79.30            & 71.32            & 27.94            & 9.21            \\
		       & \cmark & 77.14            & 68.30            & 35.04            & 11.33           \\
		\cmark & \cmark & \color{red}80.04 & \color{red}71.98 & \color{red}26.51 & \color{red}8.35 \\
		\bottomrule
	\end{tabular}
	\label{tab:abla_augs}
\end{table}

\textbf{Loss Weight Sensitivity Analysis}. To determine the optimal balance between different loss components and address how these settings influence model stability and performance, we systematically analyze the sensitivity of contrastive and consistency loss weights. As shown in \Cref{tab:abla-params}, both loss weights exhibit optimal performance at $\lambda_\mathrm{cont} = \lambda_\mathrm{consist} = 0.1$.

The contrastive loss ($\mathcal{L}_\mathrm{cont}$) is designed to make the model feature invariance to the texture and noise variations introduced by the FDS. It encourages the encoder to learn robust representations of anatomical structures by pulling features from the same spatial location in original and FDS-reconstructed images closer together. When its weight $\lambda_\mathrm{cont}$ is zero, the model loses this crucial unsupervised guidance, resulting in features that are less robust to noise and a performance drop to 77.85\% Dice. Conversely, when $\lambda_\mathrm{cont}$ is too high (\eg, 0.5), the training objective becomes dominated by instance-level discrimination rather than semantic segmentation. This objective mismatch destabilizes the learning process for the segmentation head, degrading performance to 79.54\% Dice.

Similarly, the consistency loss ($\mathcal{L}_\mathrm{consist}$) regularizes the model by enforcing similar prediction outputs for original and FDS-perturbed inputs. An optimal weight of 0.1 provides sufficient regularization and promotes model stability and generalization, while higher values impose an overly strict constraint and lead to a performance decline.

The optimal configuration achieves a crucial equilibrium. It allows the model to leverage unlabeled data to learn robust, invariant features and stable predictions, without allowing the auxiliary objectives to overpower the primary supervised segmentation task. This synergy is evidenced by the peak performance of 80.04\% Dice and a significant improvement in boundary quality (26.51 for HD95), demonstrating that carefully balanced loss weighting is critical for stable and high-performing SSL.

\begin{table}[htb]
	\caption{Ablation study of contrastive and consistency weights on the LN dataset with 5\% labeled data.}
	\centering
	\footnotesize
	\setlength{\tabcolsep}{4pt}
	\begin{tabular}{cc|cccc}
		\toprule
		Weight                       & Value & Dice$\uparrow$   & IoU$\uparrow$    & HD95$\downarrow$ & ASD$\downarrow$ \\
		\midrule
		\multirow{5}{*}{Contrastive} & 0.0   & 77.85            & 69.70            & 30.40            & 10.04           \\
		                             & 0.05  & 79.81            & 71.51            & 30.91            & 9.86            \\
		                             & 0.1   & \color{red}80.04 & \color{red}71.98 & \color{red}26.51 & \color{red}8.35 \\
		                             & 0.2   & 79.55            & 70.96            & 31.37            & 10.22           \\
		                             & 0.5   & 79.54            & 71.29            & 30.29            & 9.50            \\
		\midrule
		\multirow{5}{*}{Consistency} & 0.0   & 78.24            & 70.00            & 29.52            & 10.30           \\
		                             & 0.05  & 77.78            & 69.70            & 30.07            & 9.77            \\
		                             & 0.1   & \color{red}80.04 & \color{red}71.98 & \color{red}26.51 & \color{red}8.35 \\
		                             & 0.2   & 78.76            & 70.61            & 29.56            & 9.36            \\
		                             & 0.5   & 78.24            & 69.90            & 30.06            & 9.60            \\
		\bottomrule
	\end{tabular}
	\label{tab:abla-params}
\end{table}

\subsection{Extremely Low-labeled Scenario}

We evaluate performance in the extremely low-labeled scenario using only 1\% labeled data. The BUSI~\cite{busi_al_2020} and DDTI~\cite{ddti_pedraza_2015} datasets are excluded as they contain insufficient labeled samples to support the required batch size of 16 (8 labeled, 8 unlabeled) under this setting. Thus, we report results on the remaining four datasets.

As presented in \Cref{tab:results-3}, our method demonstrates superior robustness compared to other approaches. Specifically, on the LN-INT and LN-EXT datasets, our method outperforms the second-best ABD~\cite{abd_chi_2024} by a significant margin of 13.52\% and 18.56\% in Dice score, respectively. It also achieves the best accuracy on the Prostate~\cite{microsegnet_jiang_2023} dataset with 82.44\% Dice and 22.64 HD95. While ABD shows a slight advantage on TN3K~\cite{tn3k_gong_2023}, our method maintains competitive performance and consistently surpasses BCP~\cite{bcp_bai_2023} and $\beta$-FFT~\cite{betafft_hu_2025}. These results validate the effectiveness of our approach in extracting valuable features from unlabeled data even with minimal supervision.

\begin{table}[htb]
	\caption{Experimental results under the 1\% labeled data scenario. Numbers in parentheses refer to the quantity of labeled data.}
	\centering
	\footnotesize
	\setlength{\tabcolsep}{2pt}
	\begin{tabular}{l|cc|cc|cc|cc}
		\toprule
		\multirow{2}{*}{Method} & \multicolumn{2}{c|}{LN-INT (10)} & \multicolumn{2}{c|}{LN-EXT} & \multicolumn{2}{c|}{TN3K (23)} & \multicolumn{2}{c}{Prostate (19)}                                                                                 \\\cmidrule{2-9}
		                        & Dice$\uparrow$                   & HD95$\downarrow$            & Dice$\uparrow$                 & HD95$\downarrow$                  & Dice$\uparrow$    & HD95$\downarrow$  & Dice$\uparrow$    & HD95$\downarrow$  \\
		\midrule
		U-Net                   & 34.68                            & 79.49                       & 29.03                          & 85.52                             & 38.24             & 75.52             & 75.83             & 52.20             \\
		BCP                     & 50.49                            & 68.96                       & 39.13                          & 66.44                             & 56.80             & 46.63             & \color{blue}81.63 & \color{blue}26.47 \\
		ABD                     & \color{blue}53.10                & \color{blue}51.85           & \color{blue}41.93              & \color{blue}56.28                 & \color{red}65.07  & \color{red}36.99  & 79.80             & 26.49             \\
		$\beta$-FFT             & 52.94                            & 64.62                       & 38.88                          & 60.48                             & 57.86             & 47.15             & 78.81             & 34.53             \\
		\textbf{Switch}         & \color{red}66.62                 & \color{red}42.04            & \color{red}60.49               & \color{red}42.34                  & \color{blue}61.84 & \color{blue}37.86 & \color{red}82.44  & \color{red}22.64  \\
		\bottomrule
	\end{tabular}
	\label{tab:results-3}
\end{table}

\subsection{MSS Strategy Validation}

\begin{figure}[htb]
	\centering
	\includegraphics[width=1\linewidth]{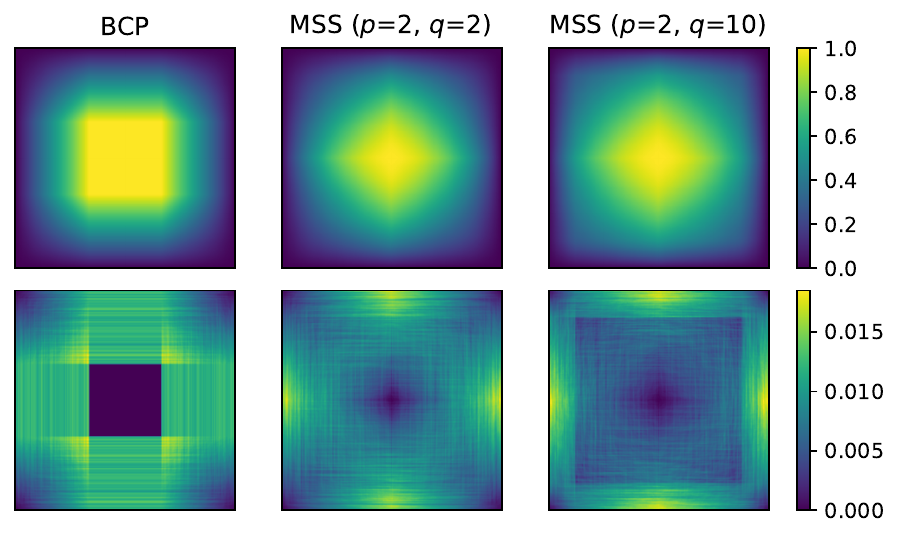}
	\caption{Quantitative comparison between BCP and MSS switching strategies. \textbf{Top}: Switching probability heatmaps (10,000 iterations) reveal geometric bias of BCP versus uniform distribution of MSS. \textbf{Bottom}: Gradient magnitude analysis shows MSS ($p$=2, $q$=2) achieves 54.86\% lower variance, indicating smoother mixing transitions, while MSS ($p$=2, $q$=10, similar area compares to BCP 2/3 strategy) also gains 8.64\% improvement. Results demonstrate MSS superiority in sample uniformity.}
	\label{fig:switch-strategy-analysis}
\end{figure}

To validate the theoretical advantages of our proposed MSS strategy, we conduct comprehensive coverage uniformity and gradient variance analysis comparing MSS with the previous SOTA BCP method~\cite{bcp_bai_2023}. \Cref{fig:switch-strategy-analysis} illustrates the quantitative comparison between MSS and BCP strategies, demonstrating the improved spatial uniformity and gradient smoothness achieved by our hierarchical multiscale approach.

Our extensive 10,000-iteration sampling analysis demonstrates that MSS ($p=2, q=2$) achieves significant quantitative improvements: 13.71\% enhancement in coverage uniformity (measured by reduced standard deviation) and 54.86\% reduction in gradient variance, indicating substantially smoother mixing transitions. To ensure fair comparison, we validated our approach against BCP with similar area constraints. As illustrated in~\Cref{fig:switch-strategy-analysis}, MSS ($p=2, q=10$) with maximum area coverage of 20/32 still demonstrates superior performance, achieving 15.41\% improvement in standard deviation and 8.64\% reduction in gradient variance compared to the BCP 2/3 strategy.

The superior performance of MSS stems from three key architectural innovations:

(1) \textit{Hierarchical Multiscale Design}. Unlike the BCP single fixed patch approach, MSS employs a coarse-to-fine hierarchy where coarse patches ($128 \times 128$) handle large anatomical structures while fine patches ($32 \times 32$) address detailed boundaries. This design eliminates the geometric bias inherent in the BCP fixed patch strategy, achieving uniform spatial coverage across different image regions and scales.

(2) \textit{Enhanced Boundary Naturalness}. The gradient variance analysis confirms that MSS produces significantly smoother transitions, indicating more natural sample mixing that preserves anatomical continuity essential for medical image analysis.

(3) \textit{Adaptive Anatomical Coverage}. The multiscale patch configuration enables adaptive coverage of irregular organ boundaries, particularly beneficial for complex morphologies in malignant US cases where traditional fixed-patch methods struggle with geometric irregularities.

\subsection{From Label Efficiency to Clinical Impact}

(1) \textit{Analysis of Label Ratio Performance Variations}. The most significant performance gains occur at 5-10\% labeling may be attributed to three complementary factors: (a) \textit{MSS adaptive coverage}: The hierarchical patch mixing strategy maximizes anatomical region coverage when ground truth annotations are limited; (b) \textit{FDS structural preservation}: Frequency domain switching maintains critical structural information while reducing reliance on pseudo-label quality, especially valuable when teacher networks have limited supervised pre-training; (c) \textit{Contrastive learning enhancement}: The CL module effectively exploits relationships within unlabeled data when supervised signals are sparse, creating robust feature representations from minimal annotations.

(2) \textit{Clinical Deployment Implications}. These performance patterns translate to significant practical advantages in clinical settings. Achieving 80\%+ Dice scores with only 5\% labeled data represents a 95\% reduction in annotation requirements while maintaining clinically acceptable segmentation quality. It enables rapid deployment across new clinical sites with minimal expert annotation effort and supports scalable implementation across multiple anatomical regions without proportional increases in annotation overhead.

\subsection{Limitations and Future Perspectives}

Despite achieving promising results, our current approach has several limitations that present opportunities for future research.

(1) \textit{Domain Specificity}. The current framework is specifically designed and optimized for US imaging with its characteristic speckle noise patterns and low-contrast boundaries. The fixed frequency area ratio ($\rho = 0.0175$) and patch sizes ($128 \times 128$ for coarse, $32 \times 32$ for fine) are empirically tuned for US characteristics. This domain-specific optimization may limit direct transferability to other medical imaging modalities such as CT and MRI. To address this limitation, adaptive parameter selection strategies and domain-agnostic frequency analysis techniques warrant investigation for enhanced cross-modality applicability.

(2) \textit{Computational Complexity}. The dual-stage training process (pre-training and self-training) coupled with FFT operations for FDS introduces additional computational burden. Exploring more efficient frequency domain operations presents a compelling research direction, including selective frequency band processing and approximated FFT implementations that maintain performance gains while reducing computational overhead.

(3) \textit{Limited Multi-Class Extension}. While our method demonstrates effectiveness for binary US image segmentation tasks, extension to multi-class scenarios introduces additional complexity in pseudo-label generation and consistency enforcement. The development of class-aware mixing strategies, hierarchical pseudo-labeling frameworks, and adaptive loss weighting mechanisms represents promising avenues for handling complex multi-organ segmentation challenges.

\section{Conclusion}
\label{sec:conclusion}

In this work, we introduced Switch, a novel semi-supervised learning framework for robust ultrasound image segmentation. Our approach synergizes Multiscale Switch to ensure uniform spatial coverage and Frequency Domain Switch to learn invariant representations against texture variations. Comprehensive experiments across six ultrasound datasets demonstrate that Switch significantly outperforms existing state-of-the-art methods, particularly in low-data regimes (1\% and 5\% labeled data). Furthermore, our framework exhibits strong cross-center generalization and boundary precision. These results validate Switch as a label-efficient and reliable solution for medical imaging, offering a promising direction for reducing annotation costs in clinical deployment.

\section{Acknowledgement}

We thank Dr Jia Ai and Dr Juan Wu from Suzhou Hospital of Traditional Chinese Medicine Affiliated to Nanjing University of Chinese Medicine for providing the LN-EXT dataset. This work was supported by General Research Funds of the Research Grant Council of Hong Kong (Reference no. 15102222 and 15102524).

\bibliographystyle{IEEEtran}
\bibliography{ref}

@String(CVPR= {IEEE Conf. Comput. Vis. Pattern Recog.})

@String(ICCV= {Int. Conf. Comput. Vis.})

@String(AAAI = {AAAI})

@String(CVPR  = {CVPR})

@String(ICCV  = {ICCV})

@article{thyroid_takashima_1995,
  author  = {Takashima, Shodayu and Fukuda, Haruki and Nomura, Naoko and Kishimoto, Haruyoshi and Kim, Tonsok and Kobayashi, Tetsuro},
  title   = {Thyroid nodules: Re-evaluation with ultrasound},
  journal = {Journal of Clinical Ultrasound},
  volume  = {23},
  number  = {3},
  pages   = {179-184},
  issn    = {0091-2751},
  doi     = {https://doi.org/10.1002/jcu.1870230306},
  year    = {1995},
  type    = {Journal Article}
}

@article{screening_brem_2015,
  author  = {Brem, Rachel F. and Lenihan, Megan J. and Lieberman, Jennifer and Torrente, Jessica},
  title   = {Screening Breast Ultrasound: Past, Present, and Future},
  journal = {American Journal of Roentgenology},
  volume  = {204},
  number  = {2},
  pages   = {234-240},
  issn    = {0361-803X},
  doi     = {10.2214/AJR.13.12072},
  year    = {2015},
  type    = {Journal Article}
}

@article{sonography_ying_2003,
  author  = {Ying, M. and Ahuja, A.},
  title   = {Sonography of Neck Lymph Nodes. Part I: Normal Lymph Nodes},
  journal = {Clinical Radiology},
  volume  = {58},
  number  = {5},
  pages   = {351-358},
  issn    = {0009-9260},
  doi     = {https://doi.org/10.1016/S0009-9260(02)00584-6},
  year    = {2003},
  type    = {Journal Article}
}

@article{sonographic_ahuja_2005,
  author  = {Ahuja, Anil T. and Ying, Michael},
  title   = {Sonographic Evaluation of Cervical Lymph Nodes},
  journal = {American Journal of Roentgenology},
  volume  = {184},
  number  = {5},
  pages   = {1691-1699},
  doi     = {10.2214/ajr.184.5.01841691},
  year    = {2005},
  type    = {Journal Article}
}

@article{ultrasound_ahuja_2008,
  author  = {Ahuja, A. T. and Ying, M. and Ho, S. Y. and Antonio, G. and Lee, Y. P. and King, A. D. and Wong, K. T.},
  title   = {Ultrasound of malignant cervical lymph nodes},
  journal = {Cancer Imaging},
  volume  = {8},
  number  = {1},
  pages   = {48-56},
  issn    = {1740-5025 (Print)
             1470-7330},
  doi     = {10.1102/1470-7330.2008.0006},
  year    = {2008},
  type    = {Journal Article}
}

@inproceedings{unet_ronneberger_2015,
  author    = {Ronneberger, Olaf and Fischer, Philipp and Brox, Thomas},
  title     = {U-net: Convolutional networks for biomedical image segmentation},
  booktitle = {Medical image computing and computer-assisted intervention–MICCAI 2015: 18th international conference, Munich, Germany, October 5-9, 2015, proceedings, part III 18},
  publisher = {Springer},
  pages     = {234-241},
  isbn      = {3319245732},
  year      = {2015},
  type      = {Conference Proceedings}
}

@article{learning_li_2020,
  title     = {Learning a convolutional neural network for propagation-based stereo image segmentation},
  author    = {Li, Xujie and Huang, Hui and Zhao, Hanli and Wang, Yandan and Hu, Mingxiao},
  journal   = {The Visual Computer},
  volume    = {36},
  number    = {1},
  pages     = {39--52},
  year      = {2020},
  publisher = {Springer}
}

@article{multilevel_zhang_2020,
  title     = {Multi-level fusion and attention-guided CNN for image dehazing},
  author    = {Zhang, Xiaoqin and Wang, Tao and Luo, Wenhan and Huang, Pengcheng},
  journal   = {IEEE Transactions on Circuits and Systems for Video Technology},
  volume    = {31},
  number    = {11},
  pages     = {4162--4173},
  year      = {2020},
  publisher = {IEEE}
}

@article{multiview_cui_2024,
  title     = {Deep multiview module adaption transfer network for subject-specific EEG recognition},
  author    = {Cui, Weigang and Xiang, Yansong and Wang, Yifan and Yu, Tao and Liao, Xiao-Feng and Hu, Bin and Li, Yang},
  journal   = {IEEE Transactions on Neural Networks and Learning Systems},
  year      = {2024},
  publisher = {IEEE}
}

@article{infrared_khan_2024,
  title     = {An infrared and visible image fusion using knowledge measures for intuitionistic fuzzy sets and Swin Transformer},
  author    = {Khan, Muhammad Jabir and Jiang, Shu and Ding, Weiping and Huang, Jiashuang and Wang, Haipeng},
  journal   = {Information Sciences},
  volume    = {684},
  pages     = {121291},
  year      = {2024},
  publisher = {Elsevier}
}

@article{retinal_zhao_2020,
  title     = {High-quality retinal vessel segmentation using generative adversarial network with a large receptive field},
  author    = {Zhao, Hanli and Qiu, Xiaqing and Lu, Wanglong and Huang, Hui and Jin, Xiaogang},
  journal   = {International Journal of Imaging Systems and Technology},
  volume    = {30},
  number    = {3},
  pages     = {828--842},
  year      = {2020},
  publisher = {Wiley Online Library}
}

@article{application_qu_2025,
  title     = {The Application of Deep Learning for Lymph Node Segmentation: A Systematic Review},
  author    = {Qu, Jingguo and Han, Xinyang and Chui, Man-Lik and Pu, Yao and Gunda, Simon Takadiyi and Chen, Ziman and Qin, Jing and King, Ann Dorothy and Chu, Winnie Chiu-Wing and Cai, Jing and others},
  journal   = {IEEE Access},
  year      = {2025},
  publisher = {IEEE}
}

@article{ai_han_2025,
  title     = {Artificial intelligence performance in ultrasound-based lymph node diagnosis: a systematic review and meta-analysis},
  author    = {Han, Xinyang and Qu, Jingguo and Chui, Man-Lik and Gunda, Simon Takadiyi and Chen, Ziman and Qin, Jing and King, Ann Dorothy and Chu, Winnie Chiu-Wing and Cai, Jing and Ying, Michael Tin-Cheung},
  journal   = {BMC cancer},
  volume    = {25},
  number    = {1},
  pages     = {73},
  year      = {2025},
  publisher = {Springer}
}

@article{adapting_qu_2025,
  title   = {Adapting vision-language foundation model for next generation medical ultrasound image analysis},
  author  = {Qu, Jingguo and Han, Xinyang and Xiao, Tonghuan and Ai, Jia and Wu, Juan and Zhao, Tong and Qin, Jing and King, Ann Dorothy and Chu, Winnie Chiu-Wing and Cai, Jing and others},
  journal = {arXiv preprint arXiv:2506.08849},
  year    = {2025}
}

@article{semi_van_2020,
  author   = {van Engelen, Jesper E. and Hoos, Holger H.},
  title    = {A survey on semi-supervised learning},
  journal  = {Machine Learning},
  volume   = {109},
  number   = {2},
  pages    = {373-440},
  issn     = {1573-0565},
  doi      = {10.1007/s10994-019-05855-6},
  url      = {https://doi.org/10.1007/s10994-019-05855-6},
  year     = {2020},
  type     = {Journal Article}
}

@article{semi_han_2020,
  author  = {Han, Luyi and Huang, Yunzhi and Dou, Haoran and Wang, Shuai and Ahamad, Sahar and Luo, Honghao and Liu, Qi and Fan, Jingfan and Zhang, Jiang},
  title   = {Semi-supervised segmentation of lesion from breast ultrasound images with attentional generative adversarial network},
  journal = {Computer methods and programs in biomedicine},
  volume  = {189},
  pages   = {105275},
  issn    = {0169-2607},
  year    = {2020},
  type    = {Journal Article}
}

@article{semi_zhai_2022,
  author  = {Zhai, Donghai and Hu, Bijie and Gong, Xun and Zou, Haipeng and Luo, Jun},
  title   = {ASS-GAN: Asymmetric semi-supervised GAN for breast ultrasound image segmentation},
  journal = {Neurocomputing},
  volume  = {493},
  pages   = {204-216},
  issn    = {0925-2312},
  year    = {2022},
  type    = {Journal Article}
}

@article{residual_farooq_2023,
  author    = {Farooq, Muhammad Umar and Ullah, Zahid and Gwak, Jeonghwan},
  journal   = {Computerized Medical Imaging and Graphics},
  year      = {2023},
  pages     = {102173},
  publisher = {Elsevier BV},
  title     = {Residual attention based uncertainty-guided mean teacher model for semi-supervised breast masses segmentation in 2D ultrasonography},
  volume    = {104}
}

@inproceedings{ph_jiang_2024,
  author    = {Jiang, Siyao and Wu, Huisi and Chen, Junyang and Zhang, Qin and Qin, Jing},
  title     = {PH-Net: Semi-Supervised Breast Lesion Segmentation via Patch-wise Hardness},
  booktitle = {Proceedings of the IEEE/CVF Conference on Computer Vision and Pattern Recognition},
  pages     = {11418-11427},
  year      = {2024},
  type      = {Conference Proceedings}
}

@inproceedings{an_wang_2019,
  author    = {Wang, Jianrong and Zhang, Ruixuan and Wei, Xi and Li, Xuewei and Yu, Mei and Zhu, Jialin and Gao, Jie and Liu, Zhiqiang and Yu, Ruiguo},
  title     = {An attention-based semi-supervised neural network for thyroid nodules segmentation},
  booktitle = {2019 IEEE International Conference on Bioinformatics and Biomedicine (BIBM)},
  publisher = {IEEE},
  pages     = {871-876},
  isbn      = {1728118670},
  year      = {2019},
  type      = {Conference Proceedings}
}

@article{deep_chen_2023,
  author  = {Chen, Fang and Chen, Lingyu and Kong, Wentao and Zhang, Weijing and Zheng, Pengfei and Sun, Liang and Zhang, Daoqiang and Liao, Hongen},
  title   = {Deep semi-supervised ultrasound image segmentation by using a shadow aware network with boundary refinement},
  journal = {IEEE Transactions on Medical Imaging},
  issn    = {0278-0062},
  year    = {2023},
  type    = {Journal Article}
}

@article{busi_al_2020,
  author  = {Al-Dhabyani, Walid and Gomaa, Mohammed and Khaled, Hussien and Fahmy, Aly},
  title   = {Dataset of breast ultrasound images},
  journal = {Data in Brief},
  volume  = {28},
  pages   = {104863},
  issn    = {2352-3409},
  doi     = {https://doi.org/10.1016/j.dib.2019.104863},
  year    = {2020},
  type    = {Journal Article}
}

@book{ddti_pedraza_2015,
  author    = {Pedraza, Lina and Vargas, Carlos and Narváez, Fabián and Durán, Oscar and Muñoz, Emma and Romero, Eduardo},
  title     = {An open access thyroid ultrasound image database},
  publisher = {SPIE},
  volume    = {9287},
  series    = {Tenth International Symposium on Medical Information Processing and Analysis},
  url       = {https://doi.org/10.1117/12.2073532},
  year      = {2015},
  type      = {Book}
}

@article{tn3k_gong_2023,
  author  = {Gong, Haifan and Chen, Jiaxin and Chen, Guanqi and Li, Haofeng and Li, Guanbin and Chen, Fei},
  title   = {Thyroid region prior guided attention for ultrasound segmentation of thyroid nodules},
  journal = {Computers in Biology and Medicine},
  volume  = {155},
  pages   = {106389},
  issn    = {0010-4825},
  doi     = {https://doi.org/10.1016/j.compbiomed.2022.106389},
  year    = {2023},
  type    = {Journal Article}
}

@article{microsegnet_jiang_2023,
  title   = {MicroSegNet: A Deep Learning Approach for Prostate Segmentation on Micro-Ultrasound Images},
  author  = {Hongxu Jiang and Muhammad Imran and Preethika Muralidharan and Anjali Patel and Jake Pensa and Muxuan Liang and Tarik Benidir and Joseph R. Grajo and Jason P Joseph and Russell Stevens Terry and John Michael DiBianco and Liyilei Su and Yuyin Zhou and Wayne Brisbane and Weimin Shao},
  journal = {Computerized medical imaging and graphics : the official journal of the Computerized Medical Imaging Society},
  year    = {2023},
  volume  = {112},
  pages   = {102326}
}

@article{vnet_milletari_2016,
  author  = {Milletarì, Fausto and Navab, Nassir and Ahmadi, Seyed-Ahmad},
  title   = {V-Net: Fully Convolutional Neural Networks for Volumetric Medical Image Segmentation},
  journal = {2016 Fourth International Conference on 3D Vision (3DV)},
  pages   = {565-571},
  year    = {2016},
  type    = {Journal Article}
}

@article{unetplusplus_zhou_2018,
  author  = {Zhou, Zongwei and Siddiquee, Md Mahfuzur Rahman and Tajbakhsh, Nima and Liang, Jianming},
  title   = {UNet++: A Nested U-Net Architecture for Medical Image Segmentation},
  journal = {Deep Learning in Medical Image Analysis and Multimodal Learning for Clinical Decision Support : 4th International Workshop, DLMIA 2018, and 8th International Workshop, ML-CDS 2018, held in conjunction with MICCAI 2018, Granada, Spain, S...},
  volume  = {11045},
  pages   = {3-11},
  year    = {2018},
  type    = {Journal Article}
}

@inproceedings{3dunet_cicek_2016,
  author    = {Çiçek, {\"O}zgün and Abdulkadir, Ahmed and Lienkamp, Soeren S. and Brox, Thomas and Ronneberger, Olaf},
  title     = {3D U-Net: Learning Dense Volumetric Segmentation from Sparse Annotation},
  booktitle = {International Conference on Medical Image Computing and Computer-Assisted Intervention},
  type      = {Conference Proceedings},
  pages     = {424-432},
  year      = {2016}
}

@article{deeplab_chen_2018,
  author  = {Chen, L. C. and Papandreou, G. and Kokkinos, I. and Murphy, K. and Yuille, A. L.},
  title   = {DeepLab: Semantic Image Segmentation with Deep Convolutional Nets, Atrous Convolution, and Fully Connected CRFs},
  journal = {IEEE Transactions on Pattern Analysis and Machine Intelligence},
  volume  = {40},
  number  = {4},
  pages   = {834-848},
  issn    = {1939-3539},
  doi     = {10.1109/TPAMI.2017.2699184},
  year    = {2018},
  type    = {Journal Article}
}

@inproceedings{pspnet_zhao_2017,
  author    = {Zhao, H. and Shi, J. and Qi, X. and Wang, X. and Jia, J.},
  title     = {Pyramid Scene Parsing Network},
  booktitle = {2017 IEEE Conference on Computer Vision and Pattern Recognition (CVPR)},
  pages     = {6230-6239},
  isbn      = {1063-6919},
  doi       = {10.1109/CVPR.2017.660},
  year      = {2017},
  type      = {Conference Proceedings}
}

@article{hrnet_wang_2021,
  author  = {Wang, J. and Sun, K. and Cheng, T. and Jiang, B. and Deng, C. and Zhao, Y. and Liu, D. and Mu, Y. and Tan, M. and Wang, X. and Liu, W. and Xiao, B.},
  title   = {Deep High-Resolution Representation Learning for Visual Recognition},
  journal = {IEEE Transactions on Pattern Analysis and Machine Intelligence},
  volume  = {43},
  number  = {10},
  pages   = {3349-3364},
  issn    = {1939-3539},
  doi     = {10.1109/TPAMI.2020.2983686},
  year    = {2021},
  type    = {Journal Article}
}

@inproceedings{meanteacher_tarvainen_2017,
  author    = {Tarvainen, Antti and Valpola, Harri},
  booktitle = {Neural Information Processing Systems},
  year      = {2017},
  pages     = {1195--1204},
  title     = {Mean teachers are better role models: Weight-averaged consistency targets improve semi-supervised deep learning results}
}

@inproceedings{cutmix_yun_2019,
  author       = {Yun, Sangdoo and Han, Dongyoon and Chun, Sanghyuk and Oh, Seong Joon and Yoo, Youngjoon and Choe, Junsuk},
  booktitle    = {IEEE International Conference on Computer Vision ({ICCV})},
  year         = {2019},
  pages        = {6022--6031},
  organization = {},
  title        = {CutMix: Regularization Strategy to Train Strong Classifiers With Localizable Features},
  volume       = {}
}

@inproceedings{copypaste_ghiasi_2021,
  title     = {Simple copy-paste is a strong data augmentation method for instance segmentation},
  author    = {Ghiasi, Golnaz and Cui, Yin and Srinivas, Aravind and Qian, Rui and Lin, Tsung-Yi and Cubuk, Ekin D and Le, Quoc V and Zoph, Barret},
  booktitle = {Proceedings of the IEEE/CVF conference on computer vision and pattern recognition},
  pages     = {2918--2928},
  year      = {2021}
}

@inproceedings{uamt_yu_2019,
  author       = {Yu, Lequan and Wang, Shujun and Li, Xiaomeng and Fu, Chi-Wing and Heng, Pheng-Ann},
  booktitle    = {International {Conference} on {Medical} {Image} {Computing} and {Computer}-{Assisted} {Intervention} ({MICCAI})},
  year         = {2019},
  pages        = {605--613},
  organization = {},
  title        = {Uncertainty-{Aware} {Self}-ensembling {Model} for {Semi}-supervised 3D {Left} {Atrium} {Segmentation}.},
  volume       = {}
}

@inproceedings{cct_ouali_2020,
  author       = {Ouali, Yassine and Hudelot, C{\' e}line and Tami, Myriam},
  booktitle    = {Computer {Vision} and {Pattern} {Recognition} ({CVPR})},
  year         = {2020},
  pages        = {12671--12681},
  organization = {},
  title        = {Semi-Supervised Semantic Segmentation With Cross-Consistency Training.},
  volume       = {}
}

@inproceedings{classmix_olsson_2021,
  author       = {Olsson, Viktor and Tranheden, Wilhelm and Pinto, Juliano and Svensson, Lennart},
  booktitle    = {IEEE/{CVF} {Winter} {Conference} on {Applications} of {Computer} {Vision} ({WACV})},
  year         = {2021},
  pages        = {1368--1377},
  organization = {},
  title        = {ClassMix: Segmentation-Based Data Augmentation for Semi-Supervised Learning},
  volume       = {}
}

@article{mcnet_wu_2022,
  author    = {Wu, Yicheng and Ge, Zongyuan and Zhang, Donghao and Xu, Minfeng and Zhang, Lei and Xia, Yong and Cai, Jianfei},
  journal   = {Medical Image Analysis},
  year      = {2022},
  pages     = {102530},
  publisher = {Elsevier BV},
  title     = {Mutual consistency learning for semi-supervised medical image segmentation},
  volume    = {81}
}

@article{ugmcl_zhang_2023,
  author    = {Zhang, Yichi and Jiao, Rushi and Liao, Qingcheng and Li, Dongyang and Zhang, Jicong},
  journal   = {Artificial Intelligence in Medicine},
  year      = {2023},
  pages     = {102476},
  publisher = {Elsevier BV},
  title     = {Uncertainty-guided mutual consistency learning for semi-supervised medical image segmentation},
  volume    = {138}
}

@inproceedings{augmentation_zhao_2023,
  author       = {Zhao, Zhen and Yang, Lihe and Long, Sifan and Pi, Jimin and Zhou, Luping and Wang, Jingdong},
  booktitle    = {Computer {Vision} and {Pattern} {Recognition} ({CVPR})},
  year         = {2023},
  pages        = {11350--11359},
  organization = {},
  title        = {Augmentation {Matters}: A {Simple}-{Yet}-{Effective} {Approach} to {Semi}-{Supervised} {Semantic} {Segmentation}.},
  volume       = {}
}

@inproceedings{bcp_bai_2023,
  author       = {Bai, Yunhao and Chen, Duowen and Li, Qingli and Shen, Wei and Wang, Yan},
  booktitle    = {Computer {Vision} and {Pattern} {Recognition} ({CVPR})},
  year         = {2023},
  pages        = {11514--11524},
  organization = {},
  title        = {Bidirectional {Copy}-{Paste} for {Semi}-{Supervised} {Medical} {Image} {Segmentation}.},
  volume       = {}
}

@inproceedings{abd_chi_2024,
  author       = {Chi, Hanyang and Pang, Jian and Zhang, Bingfeng and Liu, Weifeng},
  booktitle    = {IEEE/{CVF} {Conference} on {Computer} {Vision} and {Pattern} {Recognition} ({CVPR})},
  year         = {2024},
  pages        = {4070-4080},
  organization = {},
  title        = {Adaptive {Bidirectional} {Displacement} for {Semi}-{Supervised} {Medical} {Image} {Segmentation}},
  volume       = {}
}

@inproceedings{betafft_hu_2025,
  title     = {beta-FFT: Nonlinear Interpolation and Differentiated Training Strategies for Semi-Supervised Medical Image Segmentation},
  author    = {Hu, Ming and Yin, Jianfu and Ma, Zhuangzhuang and Ma, Jianheng and Zhu, Feiyu and Wu, Bingbing and Wen, Ya and Wu, Meng and Hu, Cong and Hu, Bingliang and others},
  booktitle = {Proceedings of the Computer Vision and Pattern Recognition Conference},
  pages     = {30839--30849},
  year      = {2025}
}

@inproceedings{curriculum_cascante_2021,
  author       = {Cascante-Bonilla, Paola and Tan, Fuwen and Qi, Yanjun and Ordonez, Vicente},
  booktitle    = {AAAI {Conference} on {Artificial} {Intelligence} ({AAAI})},
  year         = {2021},
  pages        = {6912--6920},
  organization = {},
  title        = {Curriculum {Labeling}: Revisiting {Pseudo}-{Labeling} for {Semi}-{Supervised} {Learning}.},
  volume       = {}
}

@inproceedings{stplusplus_yang_2022,
  author       = {Yang, Lihe and Zhuo, Wei and Qi, Lei and Shi, Yinghuan and Gao, Yang},
  booktitle    = {Computer {Vision} and {Pattern} {Recognition} ({CVPR})},
  year         = {2022},
  pages        = {4258--4267},
  organization = {},
  title        = {ST++: Make {Self}-{trainingWork} {Better} for {Semi}-supervised {Semantic} {Segmentation}},
  volume       = {abs/2106.05095}
}

@inproceedings{semi_wang_2022,
  author       = {Wang, Yuchao and Wang, Haochen and Shen, Yujun and Fei, Jingjing and Li, Wei and Jin, Guoqiang and Wu, Liwei and Zhao, Rui and Le, Xinyi},
  booktitle    = {Computer {Vision} and {Pattern} {Recognition} ({CVPR})},
  year         = {2022},
  pages        = {4238--4247},
  organization = {},
  title        = {Semi-{Supervised} {Semantic} {Segmentation} {Using} {Unreliable} {Pseudo}-{Labels}.},
  volume       = {}
}

@inproceedings{sam_kirillov_2023,
  author       = {Kirillov, Alexander and Mintun, Eric and Ravi, Nikhila and Mao, Hanzi and Rolland, Chlo{\' e} and Gustafson, Laura and Xiao, Tete and Whitehead, Spencer and Berg, Alexander C. and Lo, Wan-Yen and Doll{\' a}r, Piotr and Girshick, Ross B.},
  booktitle    = {IEEE {International} {Conference} on {Computer} {Vision} ({ICCV})},
  year         = {2023},
  pages        = {3992--4003},
  organization = {},
  title        = {Segment {Anything}.},
  volume       = {}
}

@article{semisam_deng_2024,
  author    = {Deng, Sen and Feng, Yidan and Lin, Haoneng and Fan, Yiting and Lee, Alex Pui-Wai and Hu, Xiaowei and Qin, Jing},
  journal   = {Proceedings of the AAAI Conference on Artificial Intelligence},
  number    = {10},
  year      = {2024},
  pages     = {11757--11765},
  publisher = {Association for the Advancement of Artificial Intelligence (AAAI)},
  title     = {Semi-supervised {TEE} {Segmentation} via {Interacting} with {SAM} {Equipped} with {Noise}-{Resilient} {Prompting}},
  volume    = {38}
}

@inproceedings{dimensionality_hadsell_2006,
  author       = {Hadsell, Raia and Chopra, Sumit and LeCun, Yann},
  booktitle    = {Computer {Vision} and {Pattern} {Recognition} ({CVPR})},
  year         = {2006},
  pages        = {1735--1742},
  organization = {},
  title        = {Dimensionality {Reduction} by {Learning} an {Invariant} {Mapping}.},
  volume       = {}
}

@inproceedings{instdisc_wu_2018,
  author       = {Wu, Zhirong and Xiong, Yuanjun and Yu, Stella X. and Lin, Dahua},
  booktitle    = {2018 {IEEE}/{CVF} {Conference} on {Computer} {Vision} and {Pattern} {Recognition}},
  year         = {2018},
  pages        = {3733-3742},
  organization = {IEEE},
  title        = {Unsupervised {Feature} {Learning} via {Non}-parametric {Instance} {Discrimination}},
  volume       = {}
}

@inproceedings{simclr_chen_2020,
  author       = {Chen, Ting and Kornblith, Simon and Norouzi, Mohammad and Hinton, Geoffrey E.},
  booktitle    = {ICML},
  year         = {2020},
  pages        = {1597-1607},
  organization = {},
  title        = {A {Simple} {Framework} for {Contrastive} {Learning} of {Visual} {Representations}},
  volume       = {}
}

@inproceedings{moco_he_2020,
  author       = {He, Kaiming and Fan, Haoqi and Wu, Yuxin and Xie, Saining and Girshick, Ross B.},
  booktitle    = {Computer {Vision} and {Pattern} {Recognition} ({CVPR})},
  year         = {2020},
  pages        = {9726--9735},
  organization = {},
  title        = {Momentum {Contrast} for {Unsupervised} {Visual} {Representation} {Learning}.},
  volume       = {}
}

@inproceedings{exploring_wang_2021,
  author       = {Wang, Wenguan and Zhou, Tianfei and Yu, Fisher and Dai, Jifeng and Konukoglu, Ender and Gool, Luc Van},
  booktitle    = {IEEE {International} {Conference} on {Computer} {Vision} ({ICCV})},
  year         = {2021},
  pages        = {7283--7293},
  organization = {},
  title        = {Exploring {Cross}-{Image} {Pixel} {Contrast} for {Semantic} {Segmentation}.},
  volume       = {}
}

@inproceedings{crosspatch_wu_2022,
  author       = {Wu, Huisi and Wang, Zhaoze and Song, Youyi and Yang, Lin and Qin, Jing},
  booktitle    = {Computer {Vision} and {Pattern} {Recognition} ({CVPR})},
  year         = {2022},
  pages        = {11656--11665},
  organization = {},
  title        = {Cross-patch {Dense} {Contrastive} {Learning} for {Semi}-supervised {Segmentation} of {Cellular} {Nuclei} in {Histopathologic} {Images}.},
  volume       = {}
}

@article{mms_lou_2023,
  author    = {Lou, Ange and Tawfik, Kareem O. and Yao, Xing and Liu, Ziteng and Noble, Jack H.},
  journal   = {IEEE Transactions on Medical Imaging},
  number    = {10},
  year      = {2023},
  pages     = {2832--2841},
  publisher = {},
  title     = {Min-{Max} {Similarity}: A {Contrastive} {Semi}-{Supervised} {Deep} {Learning} {Network} for {Surgical} {Tools} {Segmentation}.},
  volume    = {42}
}

@inproceedings{phnet_jiang_2024,
  author       = {Jiang, Siyao and Wu, Huisi and Chen, Junyang and Zhang, Qin and Qin, Jing},
  booktitle    = {IEEE/{CVF} {Conference} on {Computer} {Vision} and {Pattern} {Recognition} ({CVPR})},
  year         = {2024},
  pages        = {11418-11427},
  organization = {},
  title        = {PH-{Net}: Semi-{Supervised} {Breast} {Lesion} {Segmentation} via {Patch}-wise {Hardness}},
  volume       = {}
}

@inproceedings{swav_caron_2020,
  author       = {Caron, Mathilde and Misra, Ishan and Mairal, Julien and Goyal, Priya and Bojanowski, Piotr and Joulin, Armand},
  booktitle    = {Conference on {Neural} {Information} {Processing} {Systems} ({NeurIPS})},
  year         = {2020},
  pages        = {9912-9924},
  organization = {},
  title        = {Unsupervised {Learning} of {Visual} {Features} by {Contrasting} {Cluster} {Assignments}.},
  volume       = {}
}

@article{infonce_oord_2018,
  author    = {Oord, A{\" a}ron van den and Li, Yazhe and Vinyals, Oriol},
  journal   = {arXiv},
  year      = {2018},
  pages     = {},
  publisher = {},
  title     = {Representation {Learning} with {Contrastive} {Predictive} {Coding}},
  volume    = {abs/1807.03748}
}

@inproceedings{ssnet_wu_2022,
  author       = {Wu, Yicheng and Wu, Zhonghua and Wu, Qianyi and Ge, Zongyuan and Cai, Jianfei},
  booktitle    = {International {Conference} on {Medical} {Image} {Computing} and {Computer}-{Assisted} {Intervention} ({MICCAI})},
  year         = {2022},
  pages        = {34--43},
  organization = {},
  title        = {Exploring {Smoothness} and {Class}-{Separation} for {Semi}-supervised {Medical} {Image} {Segmentation}.},
  volume       = {}
}

@inproceedings{fda_yang_2020,
  author       = {Yang, Yanchao and Soatto, Stefano},
  booktitle    = {Computer {Vision} and {Pattern} {Recognition} ({CVPR})},
  year         = {2020},
  pages        = {4084--4094},
  organization = {},
  title        = {FDA: Fourier {Domain} {Adaptation} for {Semantic} {Segmentation}.},
  volume       = {}
}

@inproceedings{feddg_liu_2021,
  author       = {Liu, Quande and Chen, Cheng and Qin, Jing and Dou, Qi and Heng, Pheng-Ann},
  booktitle    = {Computer {Vision} and {Pattern} {Recognition} ({CVPR})},
  year         = {2021},
  pages        = {1013--1023},
  organization = {},
  title        = {FedDG: Federated {Domain} {Generalization} on {Medical} {Image} {Segmentation} via {Episodic} {Learning} in {Continuous} {Frequency} {Space}.},
  volume       = {}
}

\end{document}